%% file: main.tex
\documentclass[lettersize,journal]{IEEEtran}
\usepackage{graphicx}
\usepackage{subfig}
\usepackage{amsmath,amsfonts}
\usepackage{amsthm}
\usepackage{amssymb}
\usepackage{algorithm}
\usepackage{array}
\usepackage{textcomp}
\usepackage{stfloats}
\usepackage{url}
\usepackage{verbatim}
\usepackage{tabularx}
\usepackage{cite}
\usepackage[table,xcdraw]{xcolor}
\usepackage{booktabs}
\usepackage{multirow}
\usepackage[hidelinks]{hyperref}
\usepackage{cleveref}
\usepackage{algpseudocode}
\newtheorem{definition}{Definition}

\begin{document}

\title{GACL: Graph Attention Collaborative Learning for Temporal QoS Prediction}

\author{
Shengxiang Hu,
Guobing Zou*,
Bofeng Zhang*,
Shaogang Wu,
Shiyi Lin, 
Yanglan Gan, \\
 and Yixin Chen,~\IEEEmembership{Fellow,~IEEE
}
\IEEEcompsocitemizethanks{
    \IEEEcompsocthanksitem Shengxiang Hu, Guobing Zou, Shaogang Wu and Shiyi Lin are with the School of Computer Engineering and Science, Shanghai University, Shanghai, China. Email: \{shengxianghu, gbzou\}@shu.edu.cn.
    \IEEEcompsocthanksitem Bofeng Zhang is with the School of Computer and Information Engineering, Shanghai Polytechnic University, Shanghai, China. Email: bfzhang@sspu.edu.cn.
    \IEEEcompsocthanksitem Yanglan Gan is with the School of Computer Science and Technology, Donghua University, Shanghai 201620, China. Email: ylgan@dhu.edu.cn.
    \IEEEcompsocthanksitem Yixin Chen is with the Department of Computer Science and Engineering, Washington University in St. Louis, St. Louis, MO 63130 USA. E-mail: chen@cse.wustl.edu.
    \IEEEcompsocthanksitem * Corresponding authors}
    \thanks{}
}

\markboth{IEEE Transactions on XXX, 2024}%
{Shengxaing Hu \MakeLowercase{\textit{et al.}}: GACL: Graph Attention Collaborative Learning for Temporal QoS Prediction}


\maketitle
\begin{abstract}
Accurate prediction of temporal QoS is crucial for maintaining service reliability and enhancing user satisfaction in dynamic service-oriented environments. However, current methods often neglect high-order latent collaborative relationships and fail to dynamically adjust feature learning for specific user-service invocations, which are critical for precise feature extraction within each time slice. Moreover, the prevalent use of RNNs for modeling temporal feature evolution patterns is constrained by their inherent difficulty in managing long-range dependencies, thereby limiting the detection of long-term QoS trends across multiple time slices. These shortcomings dramatically degrade the performance of temporal QoS prediction. To address the two issues, we propose a novel \underline{G}raph \underline{A}ttention \underline{C}ollaborative \underline{L}earning (GACL) framework for temporal QoS prediction. Building on a dynamic user-service invocation graph to comprehensively model historical interactions, it designs a target-prompt graph attention network to extract deep latent features of users and services at each time slice, considering implicit target-neighboring collaborative relationships and historical QoS values. Additionally, a multi-layer Transformer encoder is introduced to uncover temporal feature evolution patterns, enhancing temporal QoS prediction. Extensive experiments on the WS-DREAM dataset demonstrate that GACL significantly outperforms state-of-the-art methods for temporal QoS prediction across multiple evaluation metrics, achieving the improvements of up to 38.80\%. 
\end{abstract}


\input{Introduction}
\input{preliminaries}
\input{approach}

\input{experiments}
\input{related_work}

\section{Conclusion and Future Work}
\label{sec:conclusion}
In this paper, we propose a novel framework for temporal QoS prediction, named \underline{\textbf{G}}raph \underline{\textbf{A}}ttention \underline{\textbf{C}}ollaborative \underline{\textbf{L}}earning (\textbf{GACL}). 
Specifically, we first leverage a dynamic user-service invocation graph to model historical interactions. Then, we design a target-prompt graph attention network to extract invocation-specific deep latent features of users and services. The target-prompt attention mechanism enhances feature extraction by considering both implicit collaborative relationships between neighbors and target users and services, and historical QoS values of corresponding user-service invocations. This dual consideration allows the network to refine initial aggregation attention scores and produce more accurate context-aware latent features for every specific user-service invocation requiring QoS prediction. Finally, the multi-layer Transformer encoder further uncovers feature temporal evolution patterns for users and services, providing a comprehensive solution for accurate QoS prediction.
Extensive experiments on the WS-DREAM dataset demonstrate GACL's superiority over state-of-the-art methods, confirming the effectiveness of our framework for accurate temporal QoS prediction.

In the future, we plan to focus on enhancing the GACL framework, which includes optimizing the architecture for various service ecosystems and integrating additional contextual information. We also aim to extend the framework to handle more complex and dynamic service environments, ensuring its adaptability and scalability for broader applications.

\section*{Acknowledgments}
This work was supported by National Natural Science Foundation of China (No. 62272290, 62172088), and Shanghai Natural Science Foundation (No. 21ZR1400400).

\bibliographystyle{IEEEtran}
\bibliography{ref}

\vspace{-20pt}
\begin{IEEEbiography}[{\includegraphics[width=1in,height=1.25in,clip,keepaspectratio]{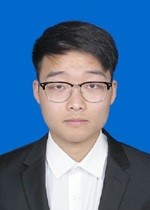}}]{Shengxiang Hu}
is currently a PhD candidate in the School of Computer Engineering and Science at Shanghai University, China. Prior to his ongoing PhD work, he successfully completed his Master's degree in Computer Science and Technology from the same university in 2021. His primary areas of research encompass Quality of Service (QoS) prediction, graph neural networks, and natural language processing. Over the course of his academic career, Hu has contributed significantly to the field through his authorship and co-authorship of 15 scholarly papers. These papers have been published in esteemed international journals and presented at prestigious conferences, such as Knowledge-based Systems, IEEE Transactions on Service Computing, ICSOC, PPSN, etc.
\end{IEEEbiography}

\vspace{-10pt}
\begin{IEEEbiography}[{\includegraphics[width=1in,height=1.3in,clip,keepaspectratio]{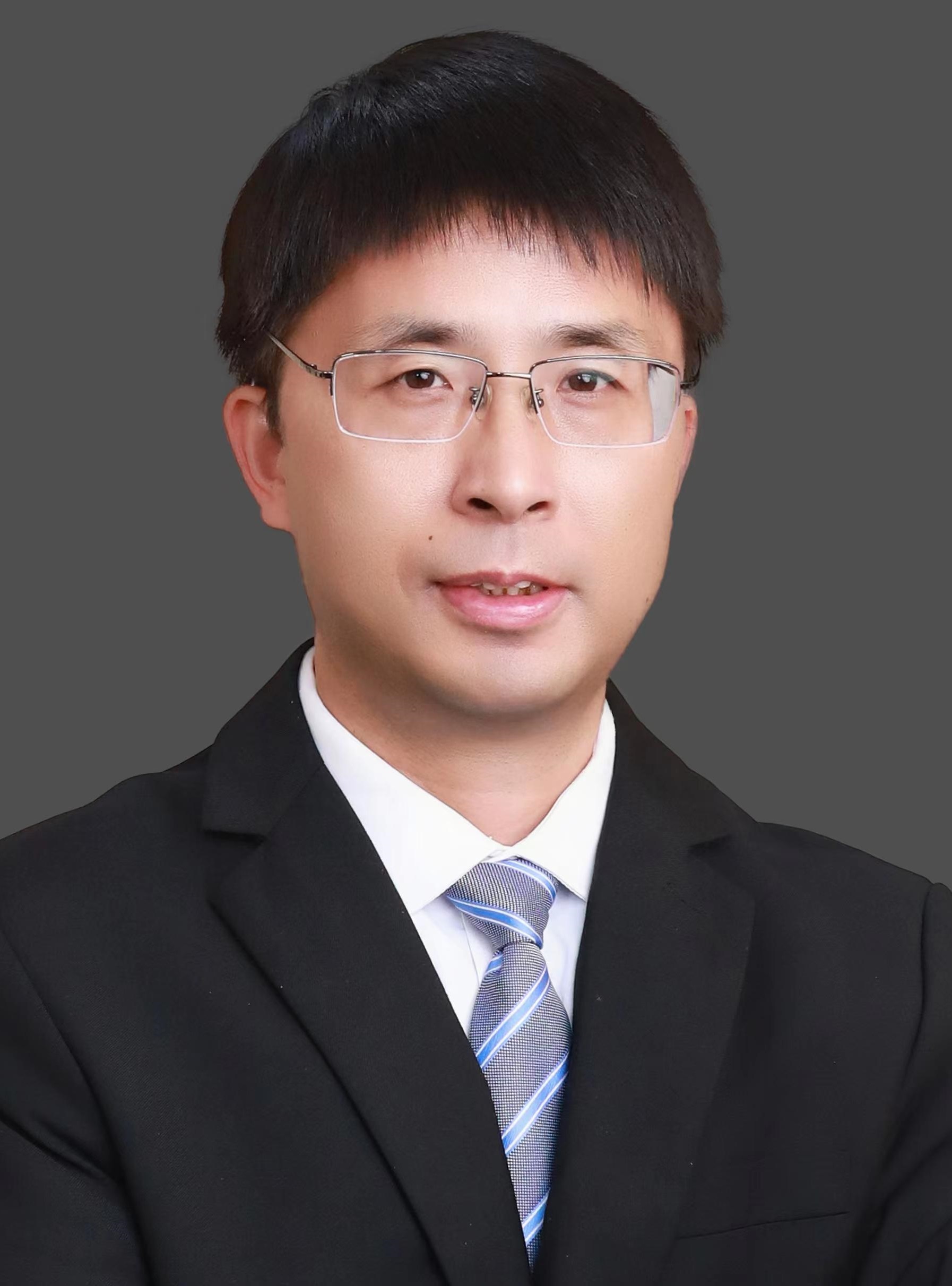}}]{Guobing Zou}
is a full professor and vice dean of the School of Computer Science, Shanghai University, China. He received his PhD degree in Computer Science from Tongji University, Shanghai, China, 2012. He has worked as a visiting scholar in the Department of Computer Science and Engineering at Washington University in St. Louis from 2009 to 2011, USA. His current research interests mainly focus on services computing, edge computing, data mining and intelligent algorithms, recommender systems.  He has published more than 120 papers on premier international journals and conferences, including IEEE Transactions on Services Computing, IEEE Transactions on Network and Service Management, IEEE ICWS, ICSOC, IEEE SCC, AAAI, Information Sciences, Expert Systems with Applications, Knowledge-Based Systems, etc.
\end{IEEEbiography}
\vspace{-30pt}

\begin{IEEEbiography}[{\includegraphics[width=1in,height=1.25in,clip,keepaspectratio]{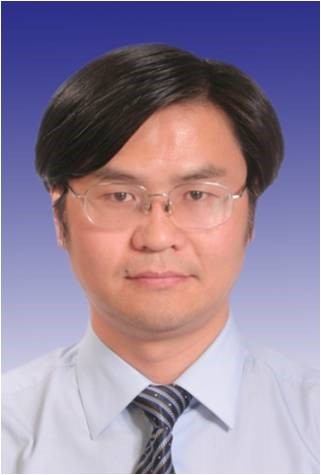}}]{Bofeng Zhang}
is a full professor and dean of the School of Computer and Information Engineering, Shanghai Polytechnic University, Shanghai, China. He received his PhD degree from the Northwestern Polytechnic University (NPU) in 1997, China. He experienced a Postdoctoral Research at Zhejiang University from 1997 to 1999, China. He worked as a visiting professor at the University of Aizu from 2006 to 2007, Japan. He worked as a visiting scholar at Purdue University from 2013 to 2014, US. His research interests include personalized service recommendation, intelligent human-computer interaction, and data mining. He has published more than 200 papers on international journals and conferences. He worked as the program chair for UUMA and ICSS. He also served as a program committee member for multiple international conferences.
\end{IEEEbiography}
\vspace{-30pt}

\begin{IEEEbiography}[{\includegraphics[width=1in,height=1.25in,clip,keepaspectratio]{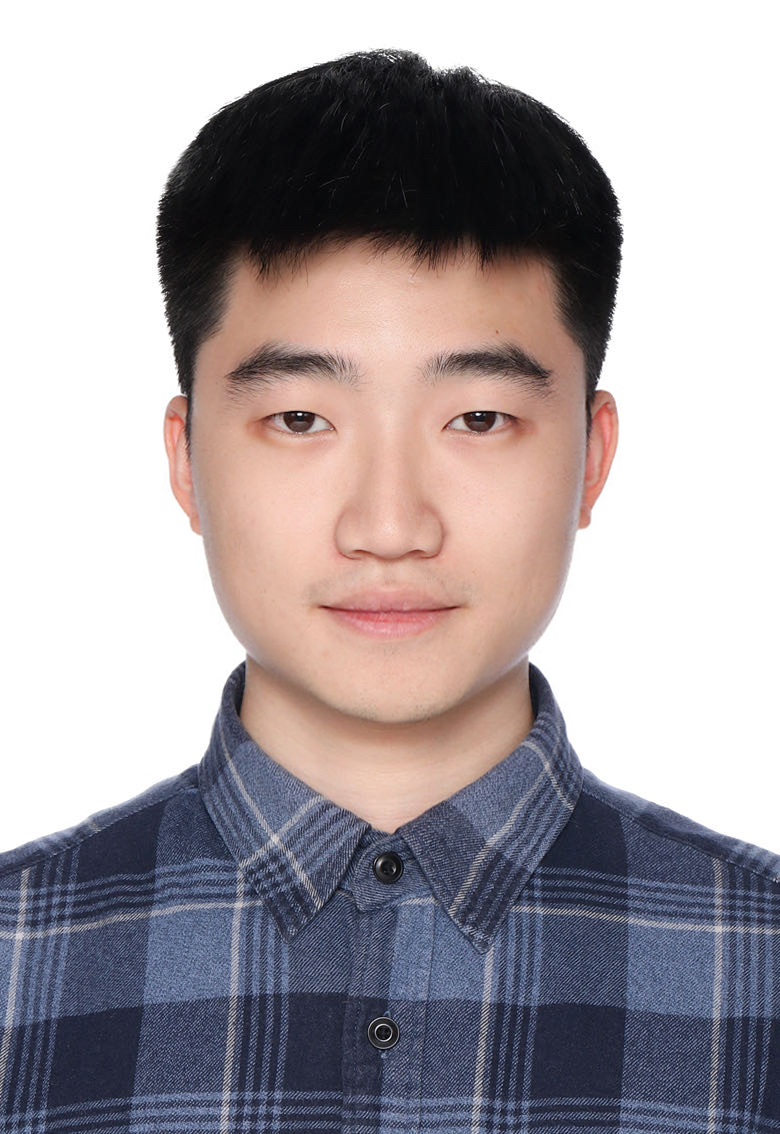}}]{Shaogang Wu}
received the bachelor’s degree in computer science and technology from Shanghai University, China in 2020. He is currently working toward the master degree in the School of Computer Engineering and Science, Shanghai University. His research interests include service quality management, deep learning, and intelligent algorithms. He has published two papers on ICSOC 2022 and TSC. He has led research and development group to successfully design and implement a service-oriented enterprise application platform, which can intelligently classify and recycle, cultivate citizens’ habit of throwing recyclables, and produce significant economic and social benefits by providing high QoS.
\end{IEEEbiography}
\vspace{-30pt}

\begin{IEEEbiography}[{\includegraphics[width=1in,height=1.25in,clip,keepaspectratio]{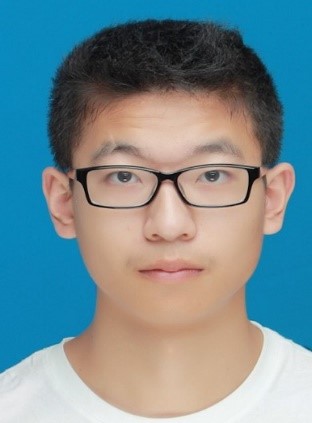}}]{Shiyi Lin}
is currently a PhD candidate in the School of Computer Engineering and Science, Shanghai University, China. He received a Bachelor degree in 2021 and Master degree in 2024 both in Computer Science and Technology at Shanghai University, respectively. His research interests include graph representation learning, recommendation systems and service computing. He has published one paper on International Conference on Service-O[tea-sdk]readyriented Computing (ICSOC), and submitted two papers on International Conference on Web Services (ICWS) and IEEE Transactions on Services Computing. 
\end{IEEEbiography}
\vspace{-30pt}

\begin{IEEEbiography}[{\includegraphics[width=1in,height=1.25in,clip,keepaspectratio]{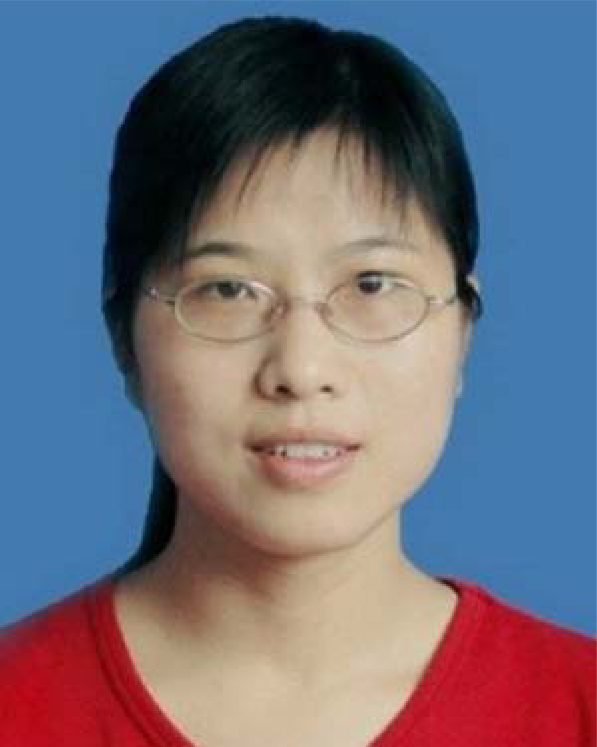}}]{Yanglan Gan}
received the PhD degree in computer science from Tongji University, Shanghai, China, 2012. She is a full professor in the School of Computer Science and Technology, Donghua University, Shanghai, China. Her research interests include bioinformatics, service computing, and data mining. She has published more than 50 papers on premier international journals and conferences, including Bioinformatics, Briefings in Bioinformatics, BMC Bioinformatics, IEEE/ACM Transactions on Computational Biology and Bioinformatics, IEEE Transactions on Services Computing, IEEE Transactions on Network and Service Management, IEEE ICWS, ICSOC, Neurocomputing, and Knowledge-Based Systems.
\end{IEEEbiography}
\vspace{-30pt}

\begin{IEEEbiography}[{\includegraphics[width=1in,height=1.25in,clip,keepaspectratio]{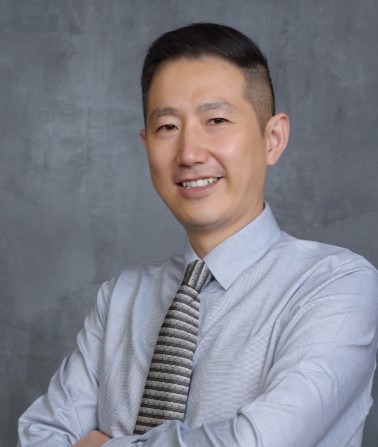}}]{Yixin Chen}
(Fellow, IEEE) received the PhD degree in computer science from the University of Illinois, Urbana Champaign, in 2005. He is currently a full professor of computer science with Washington University in St. Louis, MO, USA. His research interests include artificial intelligence, data mining, deep learning, and Big Data analytics. He has published more than 210 papers on premier international journals and conferences, including Artificial Intelligence, Journal of Artificial Intelligence Research, IEEE Transactions on Parallel and Distributed Systems, IEEE Transactions on Knowledge and Data Engineering, IEEE Transactions on Services Computing, IEEE Transactions on Computers, IEEE Transactions on Industrial Informatics, IJCAI, AAAI, ICML, KDD, etc. He won the best paper award with AAAI and a best paper nomination at KDD. He received an Early Career Principal Investigator Award from the US Department of Energy and a Microsoft Research New Faculty Fellowship. He was an associate editor for the ACM TIST, IEEE TKDE, and JAIR.
\end{IEEEbiography}

\end{document}

%% file: Introduction.tex
\section{Introduction}
\label{sec:introduction}

\IEEEPARstart{I}{n} today's interconnected service-oriented architecture, the quality of web services (QoS) is a critical metric for maintaining reliability and enhancing user satisfaction. QoS encompasses several indicators, including response time, throughput, and availability, which directly influence the overall user experience \cite{zheng2012investigating}. Faster response times enhance user satisfaction, while higher throughput and availability ensure service continuity and stability. With the global acceleration of digital transformation, the dependence of enterprises and individuals on online services has made high service quality essential \cite{kraus2022digital}. Thus, effectively managing and predicting QoS is paramount for service providers to remain competitive and consistently deliver positive user experiences.
However, accurately predicting QoS metrics remains challenging due to the dynamic and complex nature of network environments \cite{zou2022deeptsqp,liang2024qos}. Rapid changes in network traffic, server load fluctuations, and evolving user behavior patterns can significantly impact QoS, causing noticeable volatility and unpredictability. Static prediction methods \cite{shao2007personalized,wu2015collaborative,zou2018qos}, which overlook temporal trends, often fail to address these issues. Consequently, enhancing model accuracy requires a comprehensive understanding of the temporal dynamics underlying these metrics. Temporal QoS prediction aims to address these challenges by incorporating a temporal dimension that treats historical QoS data as a time-dependent sequence, thereby capturing inherent patterns and trends. Advances in machine learning and data mining techniques have led to notable progress in developing and implementing these predictive models.

Current research in temporal QoS prediction can be classified into four main categories: collaborative filtering (CF) with temporal factors \cite{hu2014time,ma2017multi,tong2021missing}, sequence prediction analysis \cite{hu2015web,ding2018time}, tensor decomposition \cite{zhang2011wspred,meng2016temporal,luo2019temporal}, and deep learning \cite{xiong2017learning,xiong2018personalized,zou2022deeptsqp}. Temporal CF selects similar neighbors using random walk algorithms and vector comparisons to address data sparsity and improve prediction accuracy. In sequence prediction analysis, the ARIMA model \cite{hu2015web,ding2018time} is adopted to enhance the prediction performance of missing temporal QoS. Tensor decomposition converts the traditional two-dimensional user-service matrix into a three-dimensional tensor, employing techniques such as CP decomposition \cite{meng2016temporal}, personalized gated recurrent units (PGRU), and generalized tensor factorization (GTF) \cite{zhang2019recurrent} comprehensively assess the impact of time. Deep learning models, such as recurrent neural networks (RNNs) and their variants like LSTM \cite{hochreiter1997long} and GRU \cite{chung2014empirical}, leverage historical QoS invocation data and multidimensional context to accurately forecast unknown QoS. By integrating temporal information, these models detect dynamic patterns and trends in QoS data, effectively addressing network volatility and complexity to enhance the precision of temporal QoS predictions.

Despite advancements in existing methods, two critical limitations still impact the performance of temporal QoS prediction.
First, although some recent works \cite{Zhang2023Predicting,Li2021Topology-Aware}, including our previous approach \cite{hu2022temporal}, have introduced neural graph learning to incorporate indirect user-service invocation relationships into the feature extraction process, thereby alleviating the sparsity problem of historical QoS invocations, they still face significant limitations in accurately capturing the unique characteristics of each user-service interaction. 
Specifically, these methods employ a uniform approach when aggregating neighbor features, failing to differentiate the importance of neighbors based on their relevance to the target user-service pair. This one-size-fits-all strategy neglects the fact that in real-world scenarios, the influence of neighboring users or services can vary greatly depending on factors such as invocation frequency, similarity in service requirements, or consistency in QoS experiences. Consequently, these models are unable to generate precise, context-aware feature representations that adapt to the specific nuances of individual QoS invocations in diverse and dynamic service environments.
Second, current models heavily rely on RNNs to capture QoS evolution across time slices. However, RNNs struggle with long-range dependencies \cite{lechner2020learning}, limiting their capacity to fully utilize historical data. This constraint hampers the ability to detect long-term QoS trends, resulting in less stable predictive performance, especially when dealing with services that exhibit complex temporal patterns or when predicting QoS over extended time horizons.

To tackle the aforementioned issues and extend our previous work, Dynamic Graph Neural Collaborative Learning (DGNCL) \cite{hu2022temporal}, we propose a novel extended framework called \underline{\textbf{G}}raph \underline{\textbf{A}}ttention \underline{\textbf{C}}ollaborative \underline{\textbf{L}}earning (\textbf{GACL}) for temporal QoS prediction.
Firstly, historical user-service QoS invocations are treated as a temporal service ecosystem and transformed into a dynamic user-service invocation graph spanning multiple consecutive time slices. It enables comprehensive modeling of the temporal evolution of user-service interactions.
Secondly, we design a target-prompt graph attention network for fine-grained, invocation-specific feature extraction of each distinct user-service pair. This network dynamically adapts to specific invocation scenarios, employing a novel target-prompt attention strategy that simultaneously leverages indirect invocation relationships and implicit collaborative correlations between targets and their neighbors. 
By recognizing the varying relevance of different neighbors and adaptively adjusting attention weights during aggregation, our model can precisely capture the unique characteristics of each user-service pair for every prediction task. 
Finally, to capture long-term temporal dependencies, we introduce a multi-layer Transformer \cite{vaswani2017attention} encoder. It effectively models the temporal evolution of the dynamically learned user and service features across extended time horizons. By uncovering complex temporal patterns, it further refines our predictions, resulting in highly accurate temporal QoS forecasts.

To evaluate the effectiveness of the proposed GACL framework, we conducted extensive experiments using two subsets of the WS-DREAM dataset \cite{zheng2012investigating}: the Response Time (RT) and Throughput (TP) datasets. Each subset consists of 4500 web services from 57 regions and 142 users from 22 regions, with a total of 27,392,643 user-service QoS invocations. These invocations are partitioned into 64 independent temporal groups of historical QoS records.
The experimental results demonstrate that GACL achieves superior performance across multiple evaluation metrics, showing significant improvements in prediction accuracy and robustness for temporal QoS prediction.

The main contributions of this paper are summarized as follows:
\begin{itemize}
  \item  We propose GACL, a novel framework for temporal QoS prediction that incorporates a dynamic user-service invocation graph to capture intricate temporal relationships, a target-prompt graph attention network to precisely learn user and service features through context-aware attention mechanisms, and a multi-layer Transformer encoder to effectively mine long-term temporal evolution patterns. By comprehensively modeling the temporal dynamics of user-service interactions, GACL achieves significant improvements in temporal QoS prediction accuracy.
  \item We design a target-prompt graph attention network that dynamically adjusts differentiated aggregation weights for each specific user-service invocation. It considers both implicit collaborative relationships between target users/services and their neighbors, as well as relevant historical QoS values, to extract more precise and invocation-specific high-order latent features within every time slice, effectively enhancing prediction accuracy in diverse and dynamic service environments.
  \item Extensive experiments are conducted on a large-scale real-world QoS dataset, and the results indicate that our proposed framework GACL receives superior performance for temporal QoS prediction compared with eight baseline approaches on MAE, NMAE and RMSE.
\end{itemize}

The remainder of this paper is structured as follows. 
Section \ref{sec:def} formulates the problem of temporal QoS prediction. Section \ref{sec:approach} elaborates the proposed framework. Section \ref{sec:experiments} presents and analyzes the experimental results. 
Section \ref{sec:related_work} reviews the related work. 
Section \ref{sec:conclusion} concludes with a discussion of future work.

%% file: preliminaries.tex
\section{Problem Definition}
\label{sec:def}


Here, we first introduce the concept of a \textit{Temporal Service Ecosystem} and provide the definition of \textit{User-Service QoS Invocation}. Then, we provide a detailed and formal definition of the \textit{Temporal QoS Prediction} problem and outline its research objectives and challenges.

\begin{definition}[Temporal Service Ecosystem]
    A temporal service ecosystem can be defined as a four-tuple $\xi=\langle \mathcal{U},\mathcal{S},\mathcal{T},\mathcal{I} \rangle$, where $\mathcal{U} = \{u_i\}_{i=1}^n$ represents a set of $n$ users, $\mathcal{S} = \{s_j\}_{j=1}^{m}$ denotes a set of $m$ web services, and $\mathcal{T} = \{t_1, t_2, \dots\}$ is a set of continuous time slices. The set $\mathcal{I} = \{\tau_{u,s}^t\}$ comprises user-service QoS invocations across multiple temporal slices.
\end{definition}

The temporal service ecosystem captures the dynamic interactions between users and services, presenting them within a temporal context to reflect interaction patterns and QoS performance across different time slices. 
A specific user-service QoS invocation in the ecosystem is defined as follows:

\begin{definition}[User-service QoS Invocation]
    Within a temporal service ecosystem $\xi$, a user-service QoS invocation is represented by a four-tuple $\tau = \langle u, s, t, r_{us}^t \rangle$, where $u \in \mathcal{U}$ denotes a user, $s \in \mathcal{S}$ is a web service, $t \in \mathcal{T}$ represents a time slice, and $r_{us}^t$ is the QoS value when $u$ invokes $s$ at $t$.
\end{definition}

Given a sequence of consecutive time slices and their corresponding user-service invocations, the problem of temporal QoS prediction can be formally defined as follows:

\begin{definition}[Temporal QoS Prediction]
    Given a temporal service ecosystem $\xi$ and the associated historical QoS matrix sequence $\mathcal{R} = \{R^t \in \mathbb{R}^{n \times m}\}_{t=1}^{|\mathcal{T}|}$, where $R^t$ is the historical QoS matrix at time slice $t$ and $r_{us}^t$ denotes the QoS value for a user $u$ and a service $s$ in $R^t$, the temporal QoS prediction problem aims to learn interaction patterns between users and services and their temporal evolution to accurately predict the QoS for user-service invocations at subsequent time slices.
    It can be formally defined as: 
    \begin{equation}
        \hat{R}^{t+1} = f(R^1, R^2, \dots, R^{|\mathcal{T}|}|\Theta_{f})
    \end{equation}
    where $\hat{R}^{t+1} \in \Re^{n\times m}$ is the predicted QoS matrix for time slice $t+1$, based on the historical QoS data $\mathcal{R}$. The function $f(\cdot | \Theta_f)$ represents the proposed prediction framework, with $\Theta_f$ being the learned parameters. 
\end{definition}
    This indicates that when a target user $u \in \mathcal{U}$ invokes a target service $s \in \mathcal{S}$ at time slice $t+1$, the predicted QoS value is $\hat{r}_{us}^{t+1} \in \hat{R}^{t+1}$.

%% file: approach.tex
\section{Approach}
\label{sec:approach}

\begin{figure*}[!t]
    \centering
    \includegraphics[width=0.95\textwidth]{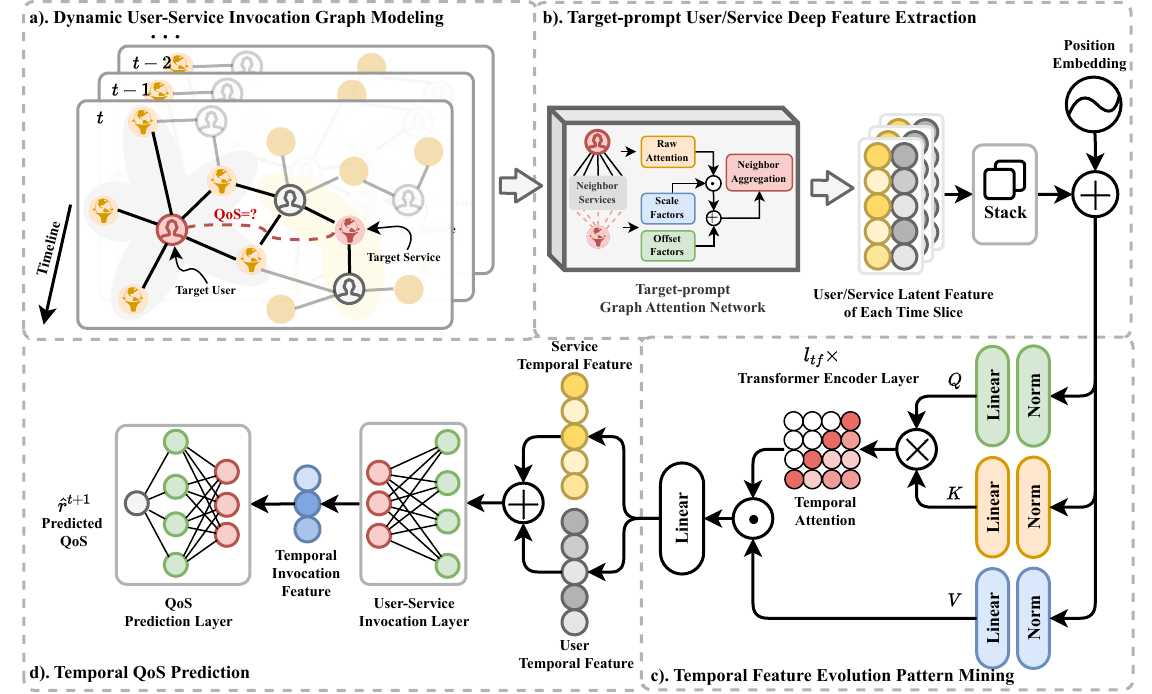}
    \caption{Overview of the GACL framework for temporal QoS prediction, which comprises four key components: a) Dynamic User-Service Invocation Graph Modeling, b) Target-prompt User/Service Deep Feature Extraction, c) Temporal Feature Evolution Pattern Mining, and d) Temporal QoS Prediction.}
    \label{fig:framework}
\end{figure*}

\begin{figure*}[t]
    \centering
    \includegraphics[width=0.9\textwidth]{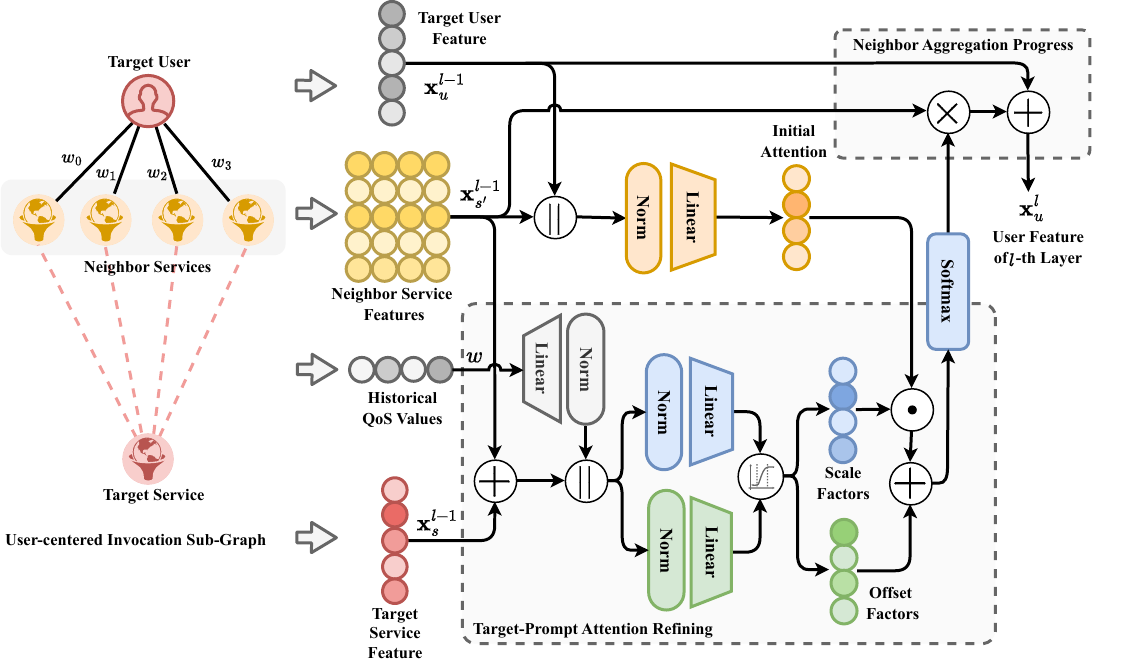}
    \caption{Message passing and aggregation in the target-prompt graph attention network. It adjusts the attention mechanism and refines the attention scores by fusing the implicit collaborative relationships between target service and the neighboring services, as well as the historical QoS values of user-service interactions. }
    \label{fig:embedding}
\end{figure*}

The overall framework of our proposed GACL is illustrated in Figure \ref{fig:framework}. It operates in the following phases:
Firstly, to represent complex historical user-service relationships, it converts a temporal service ecosystem $\xi$ with historical $|\mathcal{T}|$ time slices into a discrete dynamic user-service invocation graph. Each snapshot of this dynamic graph models user-service invocations and their corresponding QoS values for a specific time slice $t$.
Secondly, during the target-prompt user/service deep feature extraction phase, for a target user-service invocation $\tau$ whose temporal QoS needs to be predicted, the invocation graph of each time slice is sequentially input into a target-prompt graph attention network. This network extracts invocation-specific features of the user and service for each time slice by employing biased neighbor aggregation. It simultaneously considers implicit collaborative relationships between the target user/service and their neighbors, as well as the corresponding historical QoS values.
Thirdly, in the feature temporal evolution pattern mining phase, user and service features from different time slices are stacked and input into a multi-layer Transformer encoder. This encoder captures the temporal evolution dynamics of these features, generating temporal user and service features for time slice $t+1$.
Finally, two multi-layer perceptrons (MLPs) are used to extract the latent temporal user-service invocation features and predict the QoS values.

\subsection{Dynamic User-Service Invocation Graph Modeling}
Discrete dynamic graphs can effectively model complex interactions and temporal variations between entities \cite{DBLP:journals/kbs/AparicioASH24, DBLP:journals/tip/NieWHLCZ24}. Therefore, we first model the temporal service ecosystem $\xi$ with $|\mathcal{T}|$ time slices as a dynamic user-service invocation graph. Each snapshot of this dynamic graph represents user-service invocations and the corresponding QoS records for a specific time slice. In this dynamic graph, users and services are represented as nodes. Any historical user-service invocation is recorded as an edge between the target user and service in the snapshot of the appropriate time slice, with the QoS value as the edge weight. The formal definition of this dynamic invocation graph is as follows:

\begin{definition}[Dynamic User-Service Invocation Graph]
A dynamic user-service invocation graph is formulated as $\mathcal{G} = \{\mathcal{G}^t\}_{t=1}^{|\mathcal{T}|}$. For each snapshot $\mathcal{G}^t = \langle \mathcal{V}_u, \mathcal{V}_s, \mathcal{E}^t, \mathcal{W}^t \rangle$, it is a bipartite graph transformed from a sub temporal service ecosystem $\xi^t = \langle \mathcal{U}, \mathcal{S}, t, \mathcal{I}^t \rangle$ and the corresponding QoS matrix $R^t$ at time slice $t$. Here, $\mathcal{V}_u = \{v_{u_i}\}_{i=1}^n$ is a set of $n$ user vertices; $\mathcal{V}_s = \{v_{s_j}\}_{j=1}^m$ is a set of $m$ service vertices; $\mathcal{E}^t$ is a set of edges representing user-service invocation relationships. If $r^t_{u_is_j} \in R^t$, there exists an edge $e^t_{ij} = e^t_{ji} \in \mathcal{E}^t$ between $v_{u_i} \in \mathcal{V}_u$ and $v_{s_j} \in \mathcal{V}_s$; $\mathcal{W}^t$ is a set of edge weights. If $e^t_{ij} \in \mathcal{E}^t$, there exists a corresponding weight $w^t_{ij} \in \mathcal{W}^t$, which can be derived from $r^t_{u_is_j} \in R^t$.
\end{definition}

To initialize the features of each node in $\mathcal{G}$, we apply a random embedding approach. Specifically, each node $v \in \mathcal{V}_u \cup \mathcal{V}_s$ is assigned a unique ID, which is then embedded into a $d_e$-dimensional latent space to generate its initial feature, denoted as $\textbf{e}_v$. This embedding process can be formally expressed as:
\begin{equation}
\textbf{e}_v = Embedding(v)
\end{equation}
where $Embedding(\cdot)$ represents the embedding function that maps the unique ID of node $v$ to a $d_e$-dimensional vector. These embeddings serve as the initial features for the nodes in the dynamic user-service invocation graph, capturing the inherent characteristics of users and services based on their unique identifiers.

Through these steps, we construct the dynamic user-service invocation graph $\mathcal{G}$, which will be used for subsequent target-prompt deep feature learning for users and services.

\subsection{Target-prompt Deep Latent Feature Extraction of Users and Services}

Given the constructed dynamic user-service invocation graph $\mathcal{G}$ and a target invocation $\langle u, s, t+1 \rangle$ whose QoS needs to be predicted, we introduce a novel target-prompt graph attention network to extract the deep latent features of the target user $u$ and service $s$ at each previous time slice.

Building on our prior work \cite{hu2022temporal}, we recognize that a user's features are influenced by both direct invocations with services and indirect interactions with non-adjacent users and services. Similarly, a service's features are shaped by both direct user-service interactions and collaborative relationships with other services. This can be effectively achieved through aggregating neighbor information using the multi-layer recursive message passing mechanism of Graph Neural Networks (GNNs) \cite{kipf2016semi} within the invocation graph $\mathcal{G}^t$ at each time slice $t$. 
Next, we provide a detailed introduction to the feature extraction process for target users, as depicted in Figure \ref{fig:embedding}. The same methodology applies to the feature extraction for target services.

we acknowledge that neighbors contribute differently to a target user's or service's features based on contextual relevancy. Users and services with similar contexts generally indicate a more similar physical network environment. Therefore, we adopt the graph attention strategy from GAT \cite{velivckovic2017graph}, calculating the initial semantic aggregation attention of neighbor nodes based on the contextual semantic relevancy between target nodes and their neighbors. The higher the attention value, the greater the neighbor's influence on feature extraction.
Specifically, for a target user $u$, $\mathcal{N}_u^t \subseteq \mathcal{V}_s$ denotes the adjacent service vertices directly connected to $u$ in $\mathcal{G}^t$, i.e., $u$'s first-hop service neighbors at time slice $t$. For each neighbor service $s' \in \mathcal{N}_u^t$, the initial semantic aggregation attention is calculated as follows:
\begin{gather}
    (attn^t_{u\leftarrow s'})^l = \sigma_{sigmoid}(W^l_{attn}(\textbf{x}_u^{l-1} + \textbf{x}_{s'}^{l-1})^T)
\end{gather}
where $\sigma_{sigmoid}(\cdot)$ denotes the sigmoid activation function, $W^l_{attn} \in \mathbb{R}^d$ are the learnable attention parameters for the $l^{th}$ message passing layer, $\textbf{x}_u^{l-1} \in \mathbb{R}^d$ represents the hidden vector of the target user output from the $(l-1)^{th}$ layer, and $d$ is the dimensionality of the latent features for users and services. The term $(attn^t_{u \leftarrow s'})^l \in \mathbb{R}$ is the initial semantic attention value calculated at the $l^{th}$ layer. When $l=1$, this initial semantic attention is derived from the embeddings of users and services, as follows:
\begin{gather}
    \textbf{x}_v^{l-1}=\textbf{e}_v, \quad v \in \{u \cup \mathcal{N}_u^t\} \quad \text{and} \quad l=1
\end{gather}

Moreover, leveraging collaborative filtering concepts \cite{sarwar2001item,shao2007personalized}, we recognize that when predicting the QoS for various target services, those neighboring services closely related to the target service significantly contribute to the precise prediction. Thus, in extracting features for the target user, it is crucial to emphasize these similar neighboring services' features. 
Additionally, historical QoS values reflect the actual network state and the physical environment contexts between users and services. These values, represented as edge weights in the invocation graph, play a significant role in accurately capturing the interaction dynamics and should be comprehensively considered during feature aggregation. 
However, existing methods rarely consider these crucial factors simultaneously, leading to suboptimal feature learning and QoS prediction accuracy.

To mitigate this, we introduce a novel target-prompt attention strategy, depicted in the bottom right of Figure \ref{fig:embedding}, aimed at enhancing the extraction of target user features by considering the implicit collaborative relevance between neighboring and target services, as well as the corresponding historical QoS records.
Specifically, we model this enhancement as an affine transformation that adjusts the initial semantic attention based on a prompt of the target service and the historical QoS, as calculated below:
\begin{gather}
    \hat{\textbf{x}}_{s'}^{l-1} = [norm(\textbf{x}_s^{l-1} + \textbf{x}_{s'}^{l-1})||norm(W_{w}^lw_{us'}^t+b_w^l)] \label{eq:ab} \\
    (\alpha^t_{u\leftarrow s'})^l = \sigma_{tanh}(W^l_{\alpha} (\hat{\textbf{x}}_{s'}^{l-1})^T+b_{\alpha}^l) \\
    (\beta^t_{u\leftarrow s'})^l = \sigma_{tanh}(W^l_{\beta} (\hat{\textbf{x}}_{s'}^{l-1})^T+b_{\beta}^l) \\
    (\hat{attn}^t_{u\leftarrow s'})^l = (\alpha^t_{u\leftarrow s'})^l*(attn^t_{u\leftarrow s'})^l + (\beta^t_{u\leftarrow s'})^l \label{eq:attn}
\end{gather}
where $(\alpha^t_{u \leftarrow s'})^l, (\beta^t_{u \leftarrow s'})^l \in (-1, 1)$ are the learned factors that scale and shift the initial semantic attention. $\sigma_{tanh}(\cdot)$ is the hyperbolic tangent activation function, $||$ represents vector concatenation, $W_w^l \in \mathbb{R}^d$, $W^l_{\alpha}, W^l_{\beta} \in \mathbb{R}^{2d}$, and $b_w^l, b_{\alpha}^l, b_{\beta}^l$ are learnable parameters. $(\hat{attn}^t_{u \leftarrow s'})^l$ is the resulting adjusted attention value, capturing the contextual semantic relevance between the target user and neighboring services, the implicit collaborative relevance between the target service and neighboring services, as well as the influence of historical QoS values on the contribution of neighbor features.

Next, we conduct biased message propagation and aggregation based on the target-prompt attention values to enhance the feature extraction for the target user, as depicted in the top right corner of Figure \ref{fig:embedding}. 
Specifically, the message $(\textbf{m}_{u\leftarrow s'}^t)^l \in \mathbb{R}^d$, transferred from a neighboring service $s' \in \mathcal{N}_u^t$ to user $u$, is formulated as follows:
\begin{equation}
    (\textbf{m}_{u\leftarrow s'}^t)^l=\frac{exp((\hat{attn}^t_{u\leftarrow s'})^l)}{\sum_{i\in \mathcal{N}_u^t}exp((\hat{attn}^t_{u\leftarrow i})^l)}W_{msg}^l\textbf{x}_{s'}^{l-1}
\end{equation}

Here, $W_{msg}^l \in \mathbb{R}^{d \times d}$ is a trainable weight matrix of $l^{th}$ layer. The messages from all adjacent services are then aggregated as follows:
\begin{gather}
    (\textbf{x}_u^t)^l = \sigma_{prelu}((\textbf{x}_u^t)^{l-1} + \sum_{s'\in \mathcal{N}_u^t}(\textbf{m}_{u\leftarrow s'}^t)^l)    
\end{gather}
where $(\textbf{x}_u^t)^l \in \mathbb{R}^d$ represents the aggregated features of $u$, incorporating first-order messages that capture the behavioral features from directly invoked services. The function $\sigma_{prelu}$ is the PReLU \cite{he2015delving} activation function.

By stacking $l_g$ message-passing layers, we can extend the aggregation to encompass messages from $l_g$-hop user and service neighbors, culminating in the sophisticated invocation-specific high-order deep latent feature $(\textbf{x}_u^t)^{l_g}$ for user $u$. This advanced feature representation captures latent invocation correlations between $u$ and non-invoked services, as well as latent collaborative relationships among user neighbors who are structurally proximal vertices of $u$. 

Similarly, the process for learning the high-order latent feature $(\textbf{x}_s^t)^{l_g}$ for a target service $s$ mirrors that of user $u$. The distinction lies in the neighbor user message propagation and aggregation phase, where the target prompt module modifies the initial semantic attention based on the implicit collaborative relevance between neighboring users and the target user.

With a window size defined as $ws \in \mathbb{N}^{+}$, we extract deep latent features for both the target user and target service from the user-service invocation graphs over the preceding $ws$ time slices leading up to the target time slice $t+1$. This extraction yields a sequence of user features $\{(\textbf{x}_u^i)^{l_g}\}_{i=t-ws+1}^{t}$ and a corresponding sequence of service features $\{(\textbf{x}_s^i)^{l_g}\}_{i=t-ws+1}^{t}$. These feature sequences are then utilized to analyze the complex nonlinear evolutionary patterns of the target user and service over time, providing insights into their dynamic interactions and potential future behaviors.

\subsection{Temporal Feature Evolution Mining of Users and Services}
The fluctuations in the network state and invocation behavior of target users and services are reflected in the complex nonlinear evolution patterns of their features over time. To achieve more accurate temporal QoS prediction, we leverage the Transformer's \cite{vaswani2017attention} capability to comprehend long sequences, uncover these complex evolution patterns, and generate temporal features that encode these patterns for target users and services.

Specifically, to adapt to model computation, we first stack the user feature sequences and service feature sequences to obtain two feature matrices:
\begin{gather}
    X_u = stack(\{(\textbf{x}_u^i)^{l_g}\}_{i=t-ws+1}^{t}) \\
    X_s = stack(\{(\textbf{x}_s^i)^{l_g}\}_{i=t-ws+1}^{t})
\end{gather}

Next, these two feature matrices are fed into a multi-layer Transformer encoder with multi-head self-attention to explore their evolution patterns. However, the self-attention mechanism alone may not effectively identify the temporal positions of user and service features in the matrix, potentially overlooking the temporal order of the features at various time slices.
Therefore, before feeding the input into the encoder, we add positional encoding to the user and service features to explicitly incorporate their temporal nature, following the Transformer's design. For clarity, we abstract the feature matrices of the target user and target service as $X$. The specific calculation process is as follows:
\begin{gather}
    PE_{pos, 2i}=sin(\frac{pos}{10000^{2i/d}})\\
    PE_{pos, 2i+1}=cos(\frac{pos}{10000^{2i/d}}) \\
    \hat{X} = X + PE
\end{gather}
where $PE \in \mathbb{R}^{ws \times d}$ denotes the positional encoding matrix, $pos$ indexes a row in the feature matrix, and $i$ indicates the column index of one row vector.

Subsequently, we input $\hat{X}$ into a Transformer encoder with $l_{tf}$ layers and $l_{hd}$ attention heads to extract the temporal evolution patterns and generate the temporal features at $t+1$. In each encoder layer, the input first passes through the multi-head attention mechanism to capture the temporal dependencies of features at different time slices. For the $i$-th layer, the multi-head attention is calculated as follows:
\begin{gather}
    Z^0 = \hat{X} \\
    \begin{aligned}
        {Z'}^i&=MultiHead(Z^{i-1}) \\
        &=Concat(head_1^i, head_2^i,\dots,head_{l_{hd}}^i)W_{hd}^i 
    \end{aligned}
\end{gather}

Each attention head $head_j^i$ is calculated as:
\begin{gather}
    head_j^i = softmax(\frac{Q_j^i(K_j^i)^T}{\sqrt{d_k}}V_j^i) \\
    Q_j^i = Z^{i-1}(W^Q_j)^i\\
    K^i_j = Z^{i-1}(W^K_j)^i\\
    V^i_j = Z^{i-1}(W^V_j)^i
\end{gather}
where $W_j^Q$, $W_j^K$, and $W_j^V$ are the learnable weight matrices in the self-attention mechanism. $Q_j^i$ and $K_j^i$ denote the query and key matrices in self-attention, respectively. A larger attention value in a certain dimension indicates that the user and service features at that time slice contribute more significantly to generating the temporal features at $t+1$. The term $\sqrt{d_k}$ is the square root of the key vector dimension, used to scale the dot product results.

After the multi-head attention mechanism, the output passes through a feed-forward network (FFN) for nonlinear transformation. The calculation is as follows:
\begin{gather}
    \hat{Z}^i = FFN({Z'}^i) = ReLU({Z'}^iW_{1}^i+b_1^i)W_{2}^i+b_2^i \\
    Z^i = LayerNorm(\hat{Z}^i + Z^{i-1})
\end{gather}
where $W_1^i, W_2^i, b_1^i$, and $b_2^i$ are the trainable parameters of the FFN. After passing through $l_{tf}$ layers of the Transformer encoder, we obtain the temporal features of the user and service at $t+1$, encoded with the evolution dynamics:
\begin{gather}
    \textbf{h}_u^{t+1}=Z^{l_{tf}}_u[-1] \\
    \textbf{h}_s^{t+1}=Z^{l_{tf}}_s[-1]
\end{gather}

Then we take the last row of the hidden feature matrix $Z^{l_{tf}}$, which encodes the features of all $ws$ historical time slices, as the temporal features of the target user and service, $\textbf{h}_u^{t+1} \in \mathbb{R}^d$ and $\textbf{h}_s^{t+1} \in \mathbb{R}^d$, for subsequent temporal QoS prediction.

\subsection{Temporal QoS Prediction and Model Training}
Based on the target-prompt temporal features of $\textbf{h}_u^{t+1}$ and $\textbf{h}_s^{t+1}$, we concatenate them and feed the result into an MLP-based neural invocation layer to obtain the invocation feature of $u$ and $s$ at $t+1$:
\begin{gather}
    \textbf{h}_{us}^{t+1}  = \textbf{h}_u^{t+1} || \textbf{h}_s^{t+1}\\
    \textbf{h}_{inv}^{t+1}  = \sigma_{prelu}(W_{inv}\textbf{h}_{us}^{t+1}+b_{inv})\label{equ:mlp2}
\end{gather}
where $||$ denotes the concatenation operation, $W_{inv} \in \mathbb{R}^{d \times 2d}$ and $b_{inv} \in \mathbb{R}$ are the trainable parameters of the MLP.

Consequently, based on the evolutionary invocation feature $\textbf{h}_{inv}^{t+1} \in \mathbb{R}^d$, we predict the missing QoS $\hat{r}^{t+1}_{us}$ at $t+1$ using a fully connected neural network. The output layer is calculated as follows:
\begin{equation}
    \hat{r}^{t+1}_{us} = \sigma_{relu}(W_{o} \textbf{h}_{inv}^{t+1} + b_o)
\end{equation}
where $W_o \in \mathbb{R}^d$ and $b_o \in \mathbb{R}$ are the trainable output parameters, and $\hat{r}^{t+1}_{us}$ is the predicted QoS of the target invocation $\langle u, s, t+1 \rangle$.

To train and optimize the model parameters, we use the Mean Square Error (MSE) as the loss function, defined as:
\begin{equation}
    \mathcal{L} = \frac{\sum_{u \in U} \sum_{s \in S} (\hat{r}_{us}^{t+1} - r_{us}^{t+1})^2}{|U| \times |S|} + \lambda |\Theta|_2^2
\end{equation}
where $U$ and $S$ represent the user and service sets, respectively. $\Theta$ denotes all the trainable parameters of our proposed model, and $\lambda$ controls the $L2$ regularization strength to prevent overfitting. We then use mini-batch AdamW \cite{loshchilov2017decoupled} to update and optimize the parameters.

%% file: experiments.tex
\section{Experiments}
\label{sec:experiments}
\begin{table}[!t]\footnotesize 
\centering
\caption{The statistics of WS-DREAM dataset.}
\label{tab:dataset}
\begin{tabular}{@{}ccc@{}}
\toprule
\textbf{Item}             & \multicolumn{2}{c}{\textbf{Value}} \\ \midrule
Name of Sub-Dataset           & Response Time (RT)      & \multicolumn{1}{c}{Throughput (TP)}   \\
\# of Users               & \multicolumn{2}{c}{142}            \\
\# of Services            & \multicolumn{2}{c}{4,500}          \\
\# of Service Invocations & \multicolumn{2}{c}{27,392,643}     \\
Range of QoS              & 0-20s   & 0-6727kbps               \\
Mean $\&$ Variance of QoS  & 3.2$\pm$6.1 & 11.3$\pm$54.3 \\
\# of Time Slices         & \multicolumn{2}{c}{64}             \\
Overall Sparsity          & \multicolumn{2}{c}{66.98\%}        \\ \bottomrule
\end{tabular}
\end{table}

\begin{table}[!t]\footnotesize
\centering
\caption{The data excerpt of response time and throughput in WS-DREAM.}
\label{tab:data_sample}
\begin{tabular}{@{}ccccc@{}}
\toprule
\textbf{User ID} & \textbf{Service ID} & \textbf{Time Slice ID} & \textbf{RT (sec)} & \textbf{TP (kbps)}\\ \midrule
0                & 0                   & 0                      & 4.18  & 0.0868           \\
0                & 1                   & 0                      & 0.416  & 1.6370          \\
0                & 25                  & 1                      & 0.299   & 0.8523         \\ \bottomrule
\end{tabular}
\end{table}

\begin{table}[t]
\caption{Various parameter settings in the experiments.}
\label{tab:parameters}
\begin{tabularx}{0.485\textwidth}{@{}lXl@{}}
\toprule
\textbf{Parameter} & \textbf{Description}                                              & \textbf{Values}                   \\ \midrule
$l_g$     & layers of target-prompt graph attention network           & {[}1,2,3{]}              \\
$d$       & dimension of the temporal features of users and services & {[}32,64,128,256,512{]}  \\
$\rho$    & density of training dataset $\mathcal{D}_{tr}$                            & {[}5\%,10\%,15\%,20\%{]} \\
$l_{tf}$  & layers of transformer encoder                             & {[}1,2,4,8{]}            \\
$l_{hd}$  & head number of multi-head attention mechenism            & {[}1,2,4,8{]}            \\
$ws$      & window size of historical QoS records                   & {[}1,4,8,16,32{]}   \\ \bottomrule 
\end{tabularx}
\end{table}
\subsection{Experimental Setup and Dataset}
Our experimental framework was validated on a workstation equipped with an NVIDIA GTX 4090 GPU, an Intel(R) Xeon(R) Gold 6130 CPU, and 1 TB of RAM. The experimental suite for GACL was developed using Python 3.9.6, PyTorch 2.0.1, and CUDA 12.0 to ensure compatibility and optimized performance.

To evaluate the effectiveness of GACL, we conducted extensive experiments on the publicly available WS-DREAM dataset \cite{zheng2012investigating}. This large-scale real-world temporal QoS dataset has been widely used for temporal QoS prediction and includes two types of QoS criteria: response time (RT) and throughput (TP). It comprises 142 independent users, 4500 web services, and a total of 27,392,643 user-service QoS invocations for each QoS criterion, partitioned into 64 temporal groups of historical QoS records.
The overall QoS sparsity of the WS-DREAM dataset is approximately 66.98\%. 
Table \ref{tab:dataset} summarizes the detailed statistics.
To provide a more specific example, Table \ref{tab:data_sample} shows a data excerpt from WS-DREAM, where a user with ID 0 invoked a web service with ID 25 at the 1st time slice, resulting in a response time of 0.299 seconds and a throughput of 0.8523 kbps.

To align our training data with realistic application scenarios, we designed a sampling strategy that involved subsets at various densities—specifically 5\%, 10\%, 15\%, and 20\%—to form the training set $\mathcal{D}_{tr}$, while the remainder of the data was allocated to the test set $\mathcal{D}_{test}$.
Moreover, within the GACL framework, we meticulously tuned various combinations of key parameters, which are detailed in Table \ref{tab:parameters}. 
The optimal parameter combination was selected for our final experiments. 

\subsection{Competing Methods}
To evaluate the performance of our proposed GACL for temporal QoS prediction, we compare it with nine competing methods and our previous proposed one DGNCL. They are described as below:
\begin{itemize}
  \item UPCC \cite{shao2007personalized}: A user-based method using Pearson Correlation Coefficient to calculate user neighborhoods and predict QoS through deviation migration.
    \item IPCC \cite{sarwar2001item}: A service-based method that predicts QoS using service neighborhoods and combines average QoS with deviation migration.
    \item WSPred \cite{zhang2011wspred}: It extends 2D matrix factorization to 3D tensor representation for temporal QoS prediction.
    \item PNCF \cite{chen2018software}: It utilizes neural networks to learn non-linear user-service relationships from sparse vectors.
    \item PLMF \cite{xiong2018personalized}: It encodes user-service-time relationships using one-hot encoding and LSTM for temporal QoS prediction.
    \item RTF \cite{zhang2019recurrent}: It combines Personalized GRU and Generalized Tensor Factorization to analyze long-term and short-term dependency patterns.
    \item TUIPCC \cite{tong2021missing}: It integrates temporal values with CF-based QoS values using a temporal similarity computation mechanism.
    \item RNCF \cite{liang2022recurrent}: It incorporates multi-layer GRU into neural collaborative filtering to learn temporal features.
    \item GAFC \cite{10347523}: It uses probabilistic matrix factorization, gated feature extraction network, enhanced GRU, and GAN for temporal QoS prediction.
    \item DGNCL \cite{hu2022temporal}: It proposes a dynamic graph neural collaborative learning framework that combines dynamic user-service invocation graph modeling with graph convolutional networks to extract high-order latent features of users and services at each time slice, and employs a multi-layer GRU to mine temporal feature evolution patterns for temporal QoS prediction.
    \item GACL: it is our proposed approach that extends DGNCL by designing a target-prompt graph attention network for invocation-specific feature extraction. Then it introduces a Transformer encoder for temporal pattern mining. 
\end{itemize}

\begin{table*}[t]
\centering
\caption{Comparison results of various competing models on RT and TP datasets under different data density settings. Overall best results are highlighted in \textbf{bold}, and best results in competing approaches are shaded in gray.}
\label{tab:results}
\begin{tabularx}{\textwidth}{cXXXXXXXXXXXXX}
\toprule
\multirow{2}{*}{\textbf{Dataset}} & \multirow{2}{*}{\textbf{Methods}} & \multicolumn{3}{c}{\textbf{Density=5\%}} & \multicolumn{3}{c}{\textbf{Density=10\%}} & \multicolumn{3}{c}{\textbf{Density=15\%}} & \multicolumn{3}{c}{\textbf{Density=20\%}} \\ \cmidrule(l){3-14} 
&  & \textbf{MAE}  & \textbf{NMAE}  & \textbf{RMSE}  & \textbf{MAE}  & \textbf{NMAE}  & \textbf{RMSE}  & \textbf{MAE}  & \textbf{NMAE}  & \textbf{RMSE}  & \textbf{MAE}  & \textbf{NMAE}  & \textbf{RMSE}  \\ \midrule
\multirow{12}{*}{\textbf{RT}} 
& UPCC & 0.9022 & 0.2819 & 1.9243 & 0.9587 & 0.2996 & 1.7961 & 0.8948 & 0.2796 & 1.7041 & 0.8513 & 0.2660 & 1.6284 \\
& IPCC & 1.0657 & 0.3330 & 2.0001 & 0.8938 & 0.2793 & 1.7465 & 0.8432 & 0.2635 & 1.6807 & 0.8075 & 0.2523 & 1.6228 \\
& PNCF & 1.1653 & 0.3642 & 1.8358 & 1.0891 & 0.3403 & 1.7221 & 1.0427 & 0.3258 & 1.6533 & 1.0129 & 0.3165 & 1.6170 \\
& WSPred & 0.7809 & 0.2440 & 1.7065 & 0.6894 & 0.2154 & 1.6334 & 0.6726 & 0.2102 & 1.6076 & 0.6634 & 0.2073 & 1.5930 \\
& PLMF  & 0.7267 & 0.2271 & 1.7059 & 0.6786 & 0.2121 & \cellcolor[gray]{0.8}1.6126 & 0.6582 & 0.2057 & 1.5749 & 0.6444 & 0.2014 & \cellcolor[gray]{0.8}1.5523 \\
& RTF  & 0.6681 & 0.2088 & 1.7323 & 0.6302 & 0.1969 & 1.6661 & 0.6120 & 0.1912 & 1.6343 & 0.5352 & 0.1672 & 1.6361 \\
& TUIPCC  & 0.6578 & 0.2056 & 1.6484 & 0.6260 & 0.1956 & 1.6574 & 0.5749 & 0.1797 & 1.6165 & 0.5063 & 0.1582 & 1.6056 \\
& RNCF & 0.6920 & 0.2162 & 1.7582 & \cellcolor[gray]{0.8}0.6007 & \cellcolor[gray]{0.8}0.1877 & 1.6685 & 0.5902 & 0.1844 & 1.6035 & 0.5559 & 0.1737 & 1.5935 \\
& GAFC & \cellcolor[gray]{0.8}0.6318 & \cellcolor[gray]{0.8}0.1974 & \cellcolor[gray]{0.8}1.5555 & 0.6027 & 0.1883 & 1.6347 & \cellcolor[gray]{0.8}0.5511 & \cellcolor[gray]{0.8}0.1722 & \cellcolor[gray]{0.8}1.4230 & \cellcolor[gray]{0.8}0.4936 & \cellcolor[gray]{0.8}0.1542 & 1.5874 \\
\cmidrule(l){2-14}
& DGNCL & 0.5743 & 0.1795 & 1.2841 & 0.5260 & 0.1643 & 1.1934 & 0.4891 & 0.1528 & 1.1580 & 0.4618 & 0.1443 & 1.1233 \\
& GACL & \textbf{0.5126} & \textbf{0.1602} & \textbf{1.1291} & \textbf{0.4781} & \textbf{0.1328} & \textbf{1.0321} & \textbf{0.4439} & \textbf{0.1233} & \textbf{0.9907} & \textbf{0.4216} & \textbf{0.1171} & \textbf{0.9500} \\
& Gains  & 18.87\% & 18.84\% & 27.41\% & 20.41\% & 29.25\% & 36.00\% & 19.45\% & 28.40\% & 30.38\% & 14.59\% & 24.06\% & 38.80\% \\
\midrule
\multirow{12}{*}{\textbf{TP}} 
& UPCC & 4.0099 & 0.3549 & 21.9712 & 4.1034 & 0.3631 & 21.7595 & 4.1323 & 0.3657 & 21.5684 & 3.9765 & 0.3519 & 20.7731 \\
& IPCC & 4.7661 & 0.4218 & 30.3209 & 4.4244 & 0.3915 & 23.2893 & 4.4221 & 0.3913 & 23.3136 & 4.0110 & 0.3550 & 20.8456 \\
& PNCF & 4.7203 & 0.4177 & 24.1462 & 4.6581 & 0.4122 & 21.5097 & 4.5503 & 0.4027 & 20.4041 & 4.5633 & 0.4038 & 19.6659 \\
& WSPred & 4.3792 & 0.3875 & 23.6124 & 4.1666 & 0.3687 & 22.3653 & 4.1252 & 0.3651 & 22.0316 & 4.0878 & 0.3618 & 22.1613 \\
& PLMF & 4.3158 & 0.3819 & 25.6351 & 4.1234 & 0.3649 & 24.5232 & 4.0839 & 0.3614 & 23.8452 & 4.0320 & 0.3568 & 22.1021 \\
& RTF & 4.1393 & 0.3663 & 22.7817 & 4.0253 & 0.3562 & 20.8091 & 3.8740 & 0.3428 & 20.1664 & 3.7992 & 0.3362 & 19.8977 \\
& TUIPCC & 4.0752 & 0.3606 & 22.5969 & 4.0366 & 0.3572 & 20.8590 & 3.8750 & 0.3429 & 20.1663 & 3.7076 & 0.3281 & 19.0516 \\
& RNCF & 4.2877 & 0.3794 & 23.1256 & \cellcolor[gray]{0.8}3.8378 & \cellcolor[gray]{0.8}0.3396 & 20.8402 & 4.2737 & 0.3782 & 19.7895 & 3.6536 & 0.3233 & 19.3801 \\
& GAFC & \cellcolor[gray]{0.8}4.0018 & \cellcolor[gray]{0.8}0.3541 & \cellcolor[gray]{0.8}21.9519 & 3.9006 & 0.3452 & \cellcolor[gray]{0.8}19.6847 & \cellcolor[gray]{0.8}3.7405 & \cellcolor[gray]{0.8}0.3310 & \cellcolor[gray]{0.8}19.2478 & \cellcolor[gray]{0.8}3.5385 &\cellcolor[gray]{0.8} 0.3131 & \cellcolor[gray]{0.8}17.0031 \\
\cmidrule(l){2-14}
& DGNCL & 3.9824 & 0.3524 & 19.6318 & 3.9471 & 0.3493 & 18.3619 & 3.7739 & 0.3339 & 18.0271 & 3.6357 & 0.3217 & 17.5440 \\
& GACL & \textbf{3.6445} & \textbf{0.3225} & \textbf{14.1097} & \textbf{3.4006} & \textbf{0.3009} & \textbf{13.3413} & \textbf{2.8283} & \textbf{0.2503} & \textbf{12.4121} & \textbf{2.7658} & \textbf{0.2448} & \textbf{12.2824} \\
& Gains & 8.93\% & 8.92\% & 35.72\% & 11.39\% & 11.40\% & 32.23\% & 24.39\% & 24.38\% & 35.51\% & 21.84\% & 21.81\% & 27.76\% \\
\bottomrule
\end{tabularx}
\end{table*}

\subsection{Evaluation Metrics}
Given that temporal QoS prediction is essentially a regression problem, we use Mean Absolute Error (MAE), Normalized Mean Absolute Error (NMAE), and Root Mean Squared Error (RMSE) as the evaluation metrics in our experiments. These metrics quantify the deviation between the predicted QoS values and the actual ones. A smaller MAE and RMSE indicate a more accurate QoS prediction.

MAE reflects the overall accuracy of temporal QoS prediction by providing a straightforward average error magnitude. It is defined as:
\begin{equation}
MAE = \frac{\sum_{(u, s) \in \mathcal{D}} | \hat{r}_{us}^{t+1} - r_{us}^{t+1} |}{|\mathcal{D}|}
\end{equation}
where $\hat{r}_{us}^{t+1}$ signifies the predicted QoS when a target user $u$ invokes a target service $s$ at $t+1$, $r_{us}^{t+1}$ represents the actual QoS value, $\mathcal{D}$ denotes the set of QoS samples to be predicted, and $|\mathcal{D}|$ is the number of QoS samples in the dataset.

NMAE provides a relative measure of the error by normalizing the MAE against the sum of actual QoS values. This normalization helps in comparing the prediction errors across different datasets or scenarios with varying QoS scales. It is defined as:
\begin{equation}
NMAE = \frac{MAE \cdot |\mathcal{D}|}{\sum_{(u, s) \in \mathcal{D}} r_{us}^{t+1}}
\end{equation}

RMSE gives higher weight to larger errors due to its squared term, making it more sensitive to outliers in the predicted QoS values. This sensitivity to outliers is crucial for identifying significant deviations and ensuring robust performance across diverse prediction scenarios. It is defined as:
\begin{equation}
RMSE = \sqrt{\frac{\sum_{(u, s) \in \mathcal{D}} (\hat{r}_{us}^{t+1} - r_{us}^{t+1})^2}{|\mathcal{D}|}}
\end{equation}

In our experiments, MAE, NAME and RMSE are utilized to comprehensively assess the performance of our temporal QoS prediction model. MAE provides an average error magnitude, NMAE offers a normalized perspective of the prediction error, and RMSE highlights the impact of larger errors. By employing these metrics, we ensure both overall accuracy and robustness against outliers in our evaluations.

\subsection{Competing Results and Analyses}
Table \ref{tab:results} presents the comparative experimental results of our proposed GACL framework and various baseline methods on the RT and TP datasets under different data density settings (5\%, 10\%, 15\%, 20\%). The best results are highlighted in bold, while the second-best results are shaded in gray.

As shown in the table, our proposed GACL consistently achieves the best performance on both the RT and TP datasets. 
Specifically, on the RT dataset, GACL achieved improvements ranging from 14.59\% (MAE improvement compared to GAFC at a density of 20\%) to 38.80\% (RMSE improvement compared to PLMF at a density of 20\%) over the best-performing baseline across different metrics. On the TP dataset, GACL achieved improvements ranging from 8.92\% (NMAE improvement compared to GAFC at a density of 5\%) to 35.72\% (RMSE improvement compared to GAFC at a density of 5\%) over the best-performing baseline across different metrics.
Additionally, from the table, it can be seen that GACL's improvement in the RMSE metric (an average of 33.15\% in RT and 32.81\% in TP) is generally more significant than in MAE and NMAE (an average of 10.04\% in RT and 16.64\% in TP). This is because RMSE is sensitive to outliers in the dataset, indicating that compared to baselines, GACL has better robustness and generalization ability.

The significant performance enhancement demonstrates that our proposed target-prompt GAT effectively considers the complex implicit collaborative correlations and indirect invocation relationships between targets and their neighbors for every specific invocation. This allows GACL to extract the invocation-specific features of target users and services in each time slice and leverage the Transformer's powerful sequence modeling capabilities to learn the temporal evolution patterns of user and service features, achieving precise temporal QoS prediction.

Regarding the performance changes under different data density settings, we observe that the performance of GACL consistently outperforms the baselines. This indicates that by designing a target-prompt GAT and adopting multi-hop biased message passing, GACL effectively aggregates information from both directly and indirectly interacting neighbors to enhance the feature representation of target users and services, thereby alleviating the problem of accurately predicting QoS in sparse scenarios. Additionally, as data density increases, the performance of all models improves; however, the gain in performance for GACL is more significant compared to the baselines. This is reflected in how GACL's relative gains expand with increased data density, demonstrating that GACL can more effectively extract valuable information from increasing data to achieve accurate temporal QoS prediction and integrate it into high-order temporal latent features for users and services.

Moreover, as shown in the table, GACL achieves significant improvements in QoS prediction performance across all experimental settings compared to our previous work, DGNCL. It demonstrates that the target-prompt GAT in GACL effectively identifies similar neighboring users and services under the guidance of target prompts. By simultaneously integrating historical QoS records, it facilitates high-quality deep feature extraction for users and services. Additionally, GACL enhances temporal feature extraction by replacing DGNCL's GRU with a Transformer encoder. The Transformer's self-attention mechanism enables comprehensive information aggregation across all time steps, addressing GRU's limitations in processing long-range dependencies. This architectural advancement allows GACL to more effectively capture global patterns and multi-scale temporal dynamics in QoS data, resulting in more accurate 

Overall, our proposed GACL achieves state-of-the-art performance on both the RT and TP datasets. Its excellent performance across different QoS density settings validates its effectiveness in handling QoS sparsity and capturing temporal features. The proposed target-prompt GAT and multi-layer Transformer encoder show significant advantages in high-order latent feature extraction and temporal evolution pattern mining for every specific target invocation, providing the robustness for precise temporal QoS prediction.

\begin{figure*}[t]
    \centering
    \subfloat[NMAE of ablated variants on RT]{%
        \includegraphics[width=0.23\textwidth]{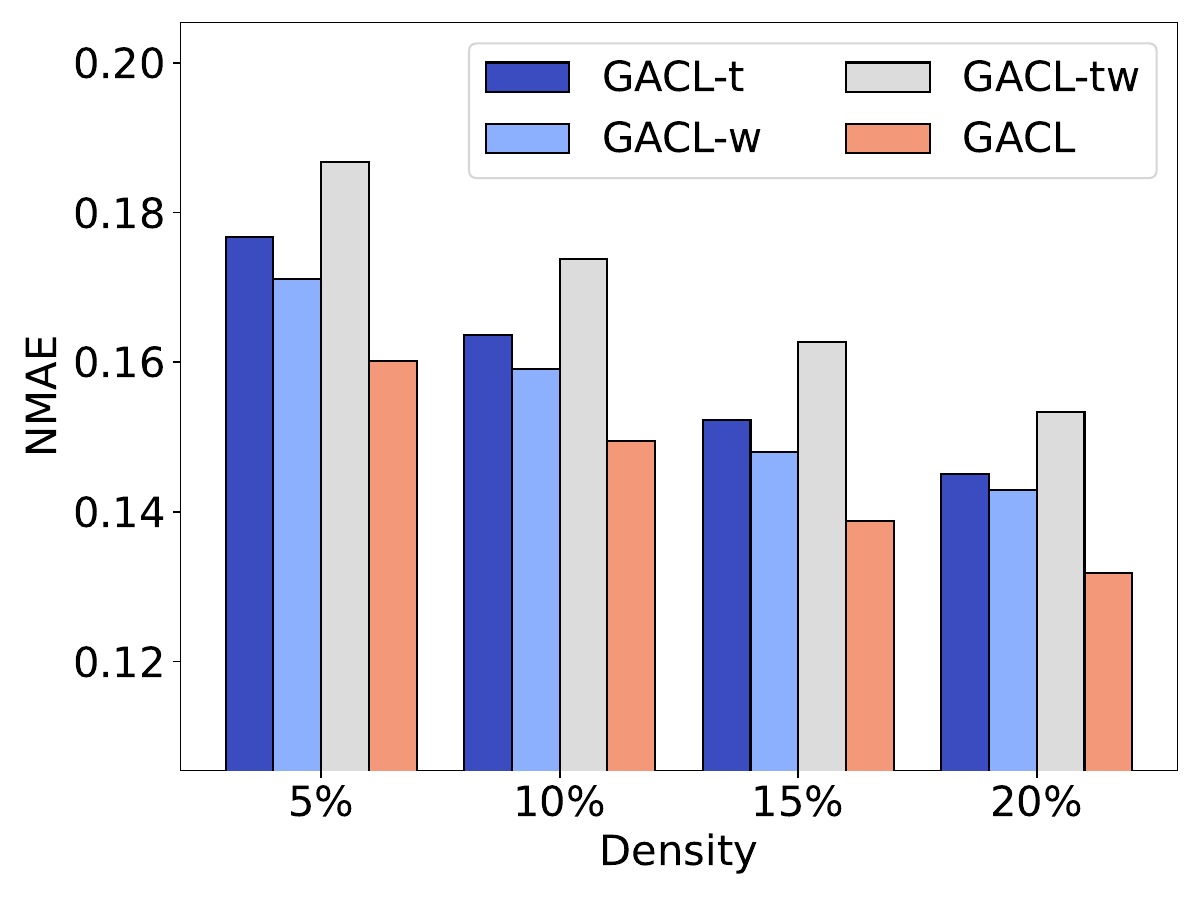}
        \label{fig:ab_rt_mae}
    }
    \hfill
    \subfloat[RMSE of ablated variants on RT]{%
        \includegraphics[width=0.23\textwidth]{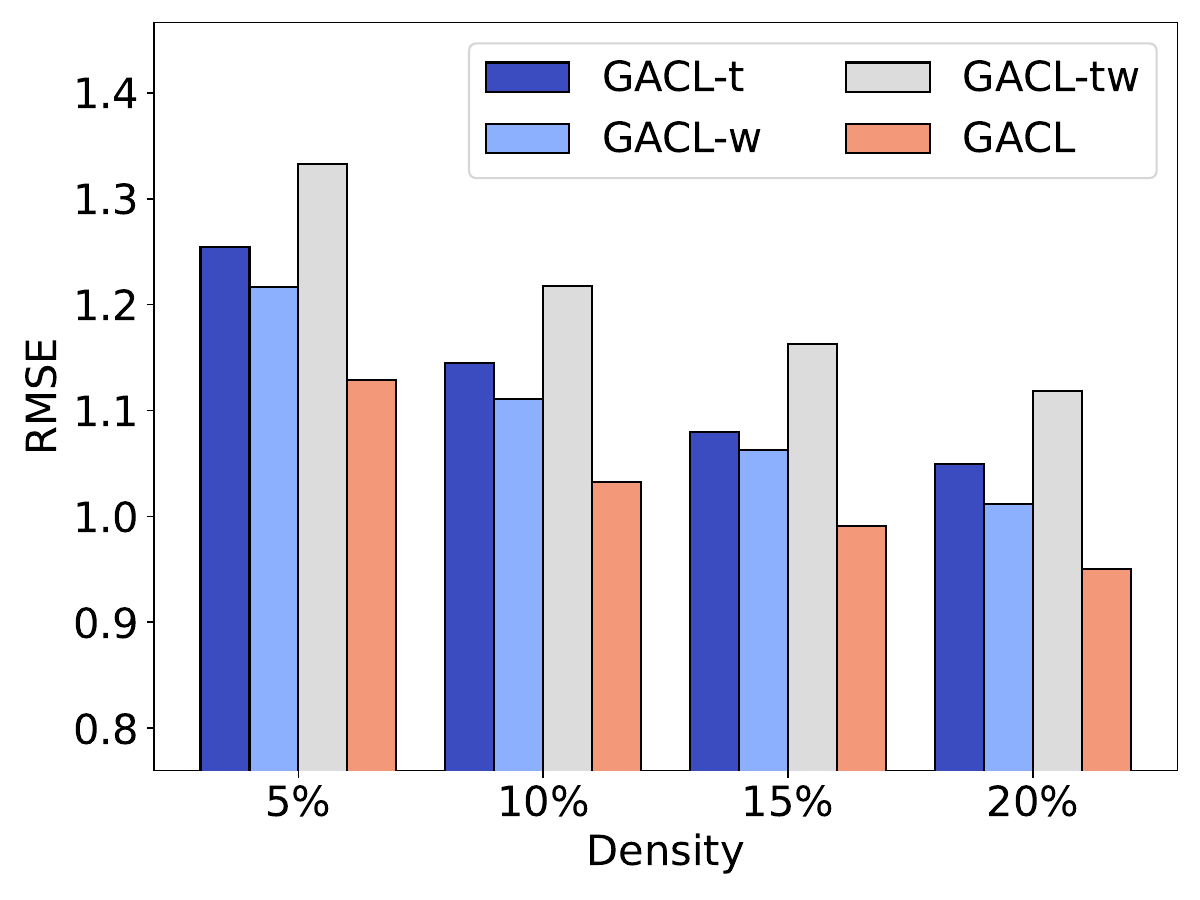}
        \label{fig:ab_rt_rmse}
    }
    \hfill
    \subfloat[NMAE of ablated variants on TP]{%
        \includegraphics[width=0.23\textwidth]{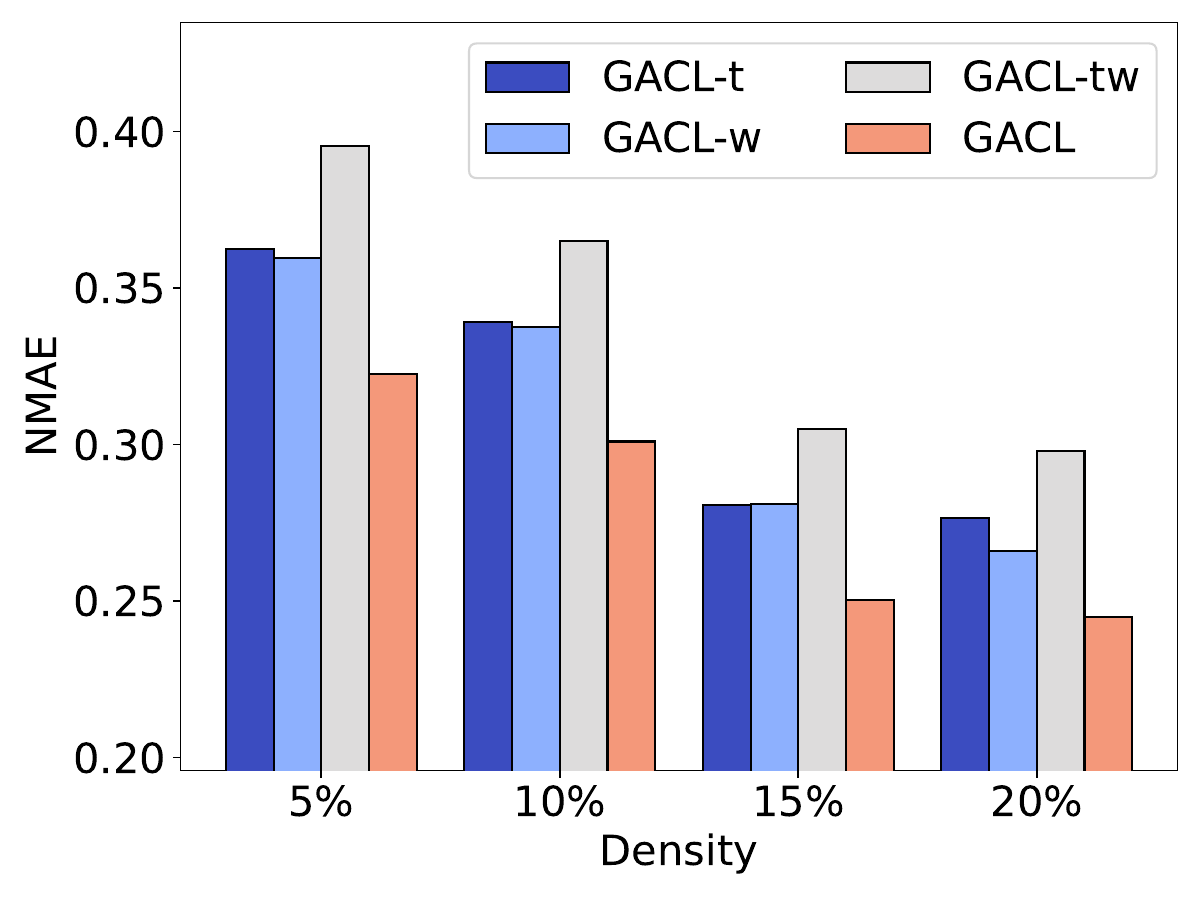}
        \label{fig:ab_tp_mae}
    }
    \hfill
    \subfloat[RMSE of ablated variants on TP]{%
        \includegraphics[width=0.23\textwidth]{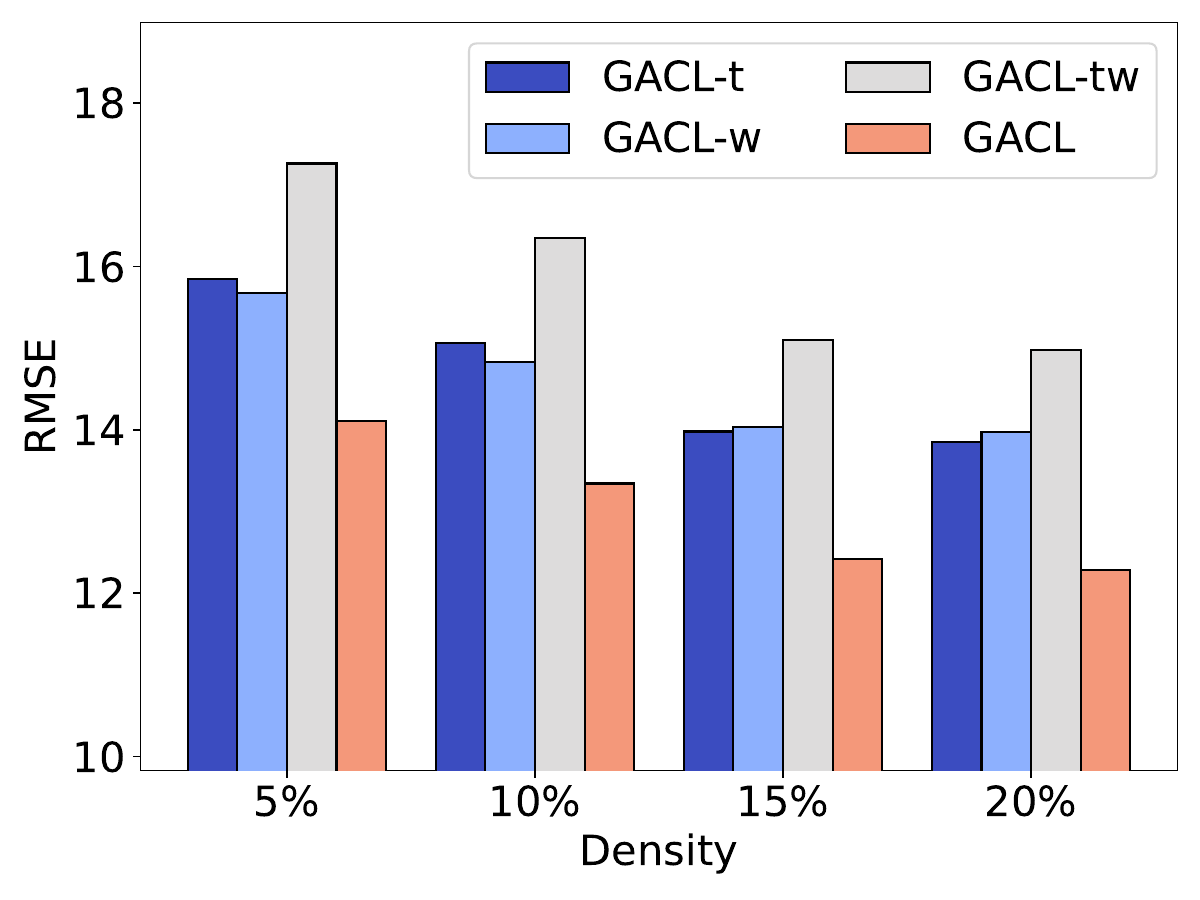}
        \label{fig:ab_tp_rmse}
    }
    \caption{Ablation experiment results of NMAE and RMSE over RT and TP.}
    \label{fig:ablations}
\end{figure*}

\subsection{Ablation Study}

To validate the effectiveness of the proposed target-prompt attention strategy, which aims to aggregate neighbor features while considering the implicit collaborative relationships between neighbors and both target user and service, as well as the historical QoS values of user-service invocations, we designed a series of ablation experiments. These experiments were conducted on both the RT and TP datasets, reporting the NMAE and RMSE metrics.
In our experiments, we conducted three ablation variants:
\begin{itemize}
    \item GACL-t: It does not consider the implicit collaborative relationships between the target service and neighbor services (or between the target user and neighbor users) during feature aggregation. In this case, \cref{eq:ab} is modified as follows:
    \begin{equation}
        \hat{\textbf{x}}_{s'}^{l-1} = norm(W_{w}^lw_{us'}^t+b_w^l)
    \end{equation}
    \item GACL-w: It does not consider the historical QoS values. Thus, \cref{eq:ab} is changed to:
    \begin{equation}
        \hat{\textbf{x}}_{s'}^{l-1} = norm(\textbf{x}_s^{l-1} + \textbf{x}_{s'}^{l-1})
    \end{equation}
    \item GACL-tw: It does not consider either the implicit collaborative relationships or the historical QoS values, relying only on the semantic relevance between the target and its 1-hop neighbors. Therefore, \cref{eq:attn} is modified as:
    \begin{equation}
        (\hat{attn}^t_{u\leftarrow s'})^l = (attn^t_{u\leftarrow s'})^l
    \end{equation}
\end{itemize}

\begin{figure*}[t]
    \centering
    \subfloat[NMAE of various $l_g$ on RT]{%
        \includegraphics[width=0.23\textwidth]{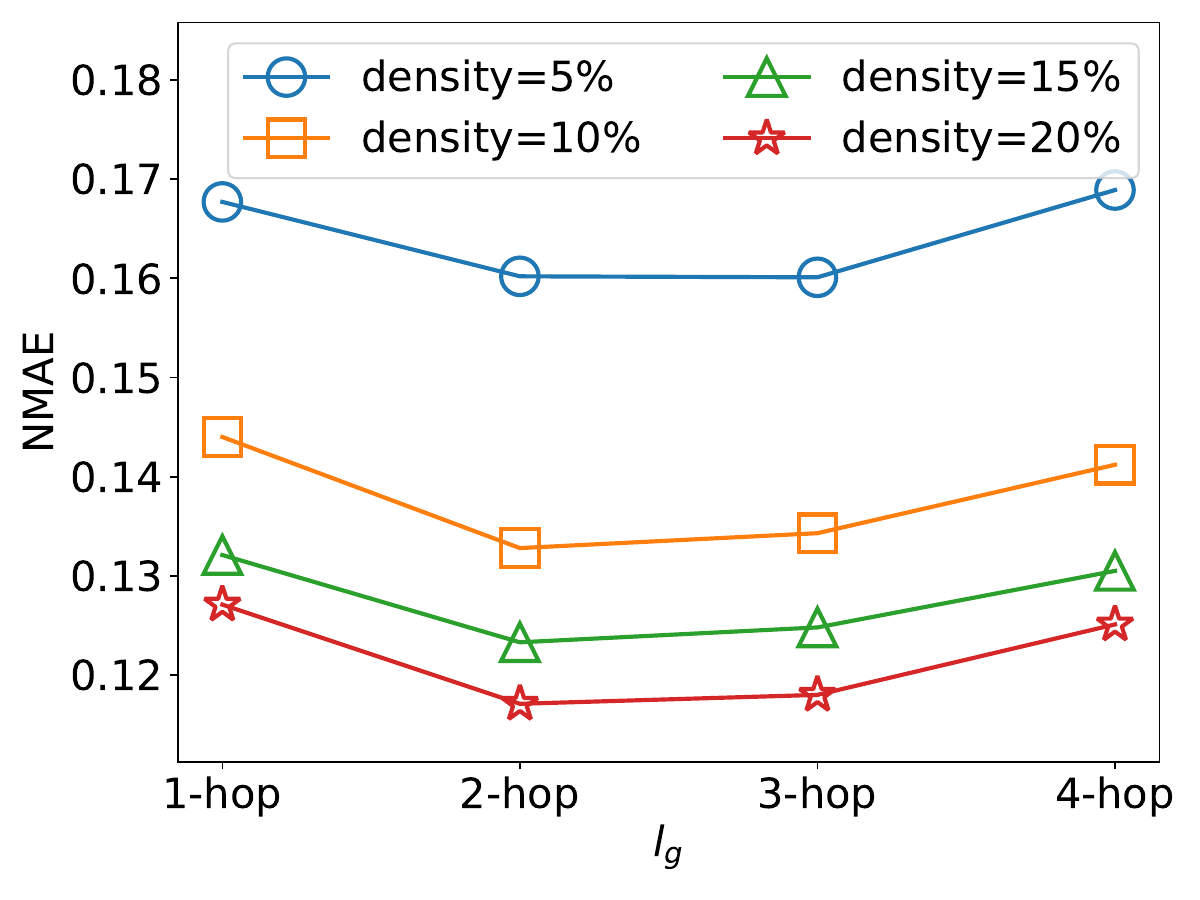}
        \label{fig:hop_rt_mae}
    }
    \hfill
    \subfloat[RMSE of various $l_g$ on RT]{%
        \includegraphics[width=0.23\textwidth]{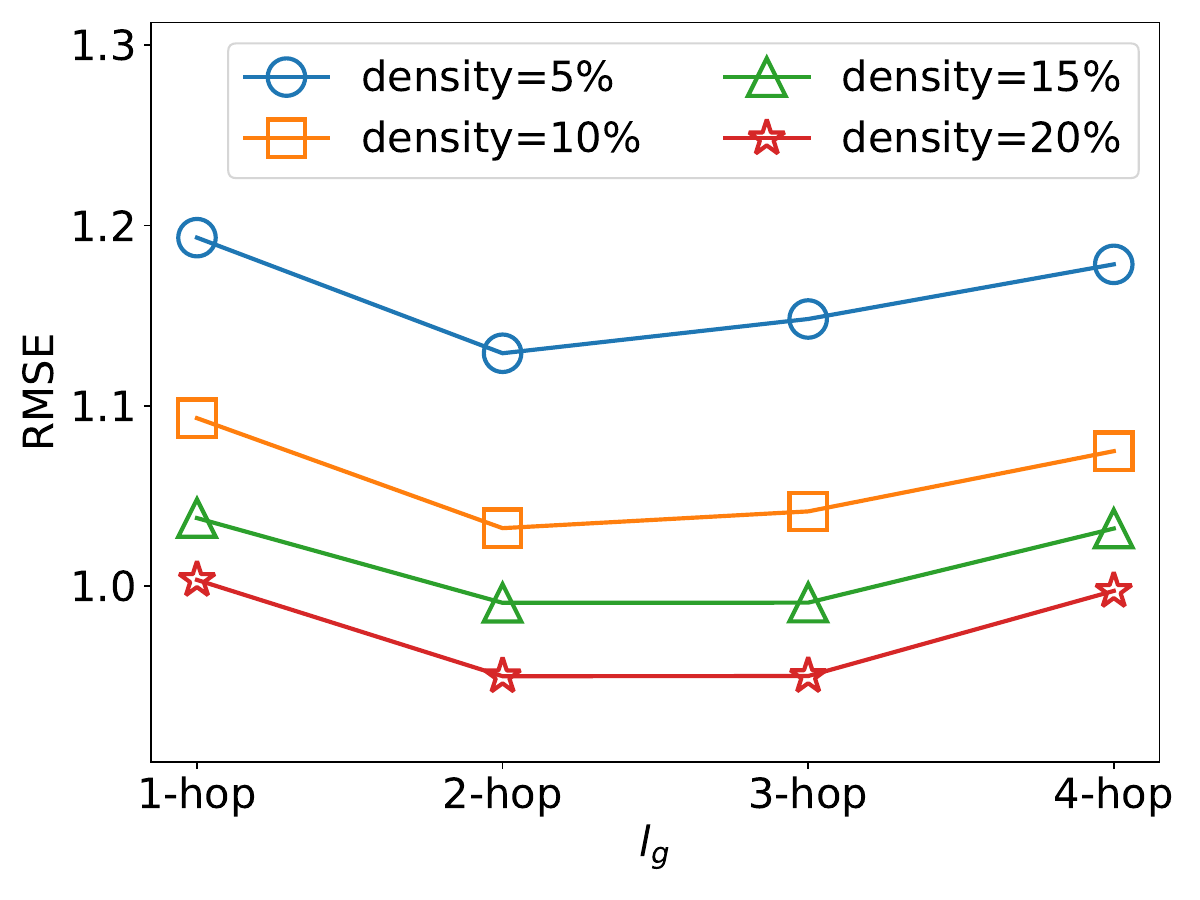}
        \label{fig:hop_rt_rmse}
    }
    \hfill
    \subfloat[NMAE of various $l_g$ on TP]{%
        \includegraphics[width=0.23\textwidth]{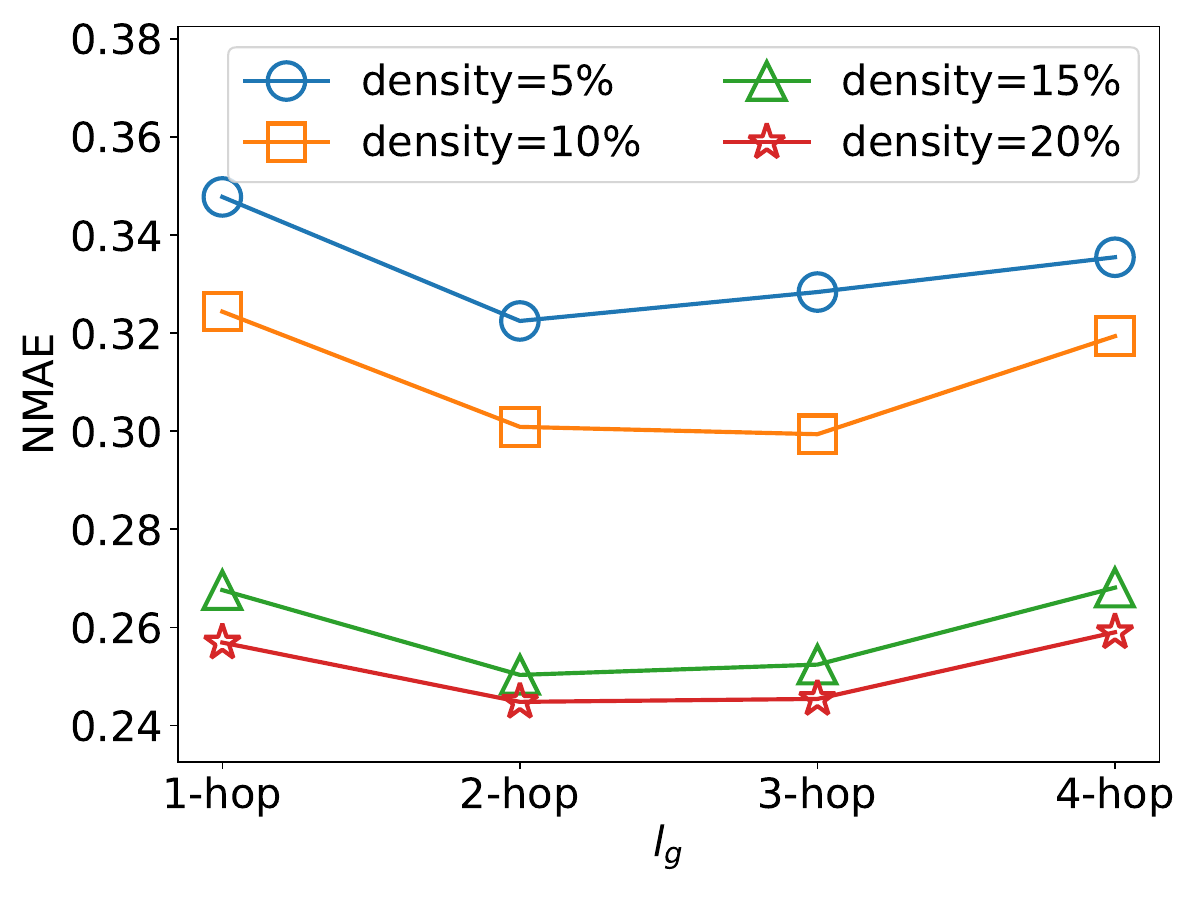}
        \label{fig:hop_tp_mae}
    }
    \hfill
    \subfloat[RMSE of various $l_g$ on TP]{%
        \includegraphics[width=0.23\textwidth]{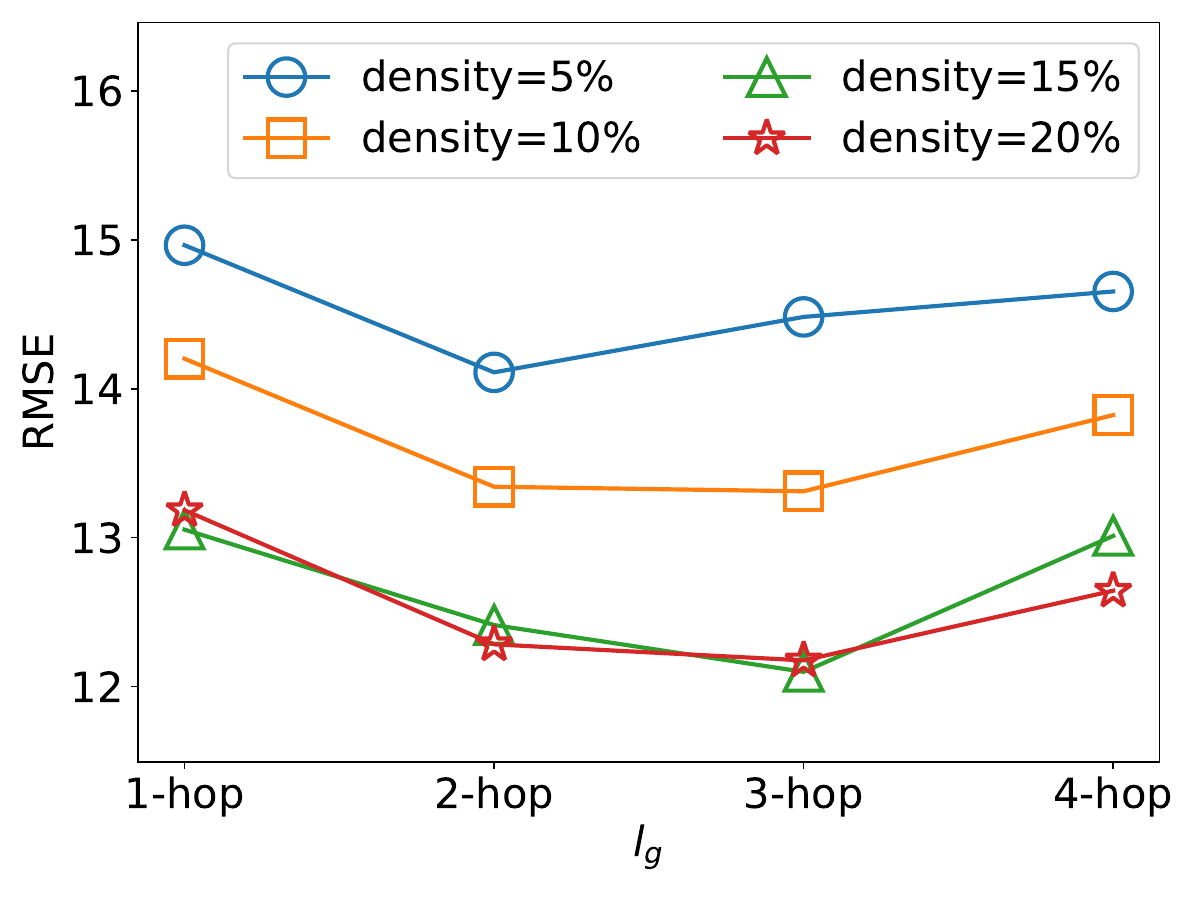}
        \label{fig:hop_tp_rmse}
    }

    \centering
    \subfloat[NMAE of various $d$ on RT]{%
        \includegraphics[width=0.23\textwidth]{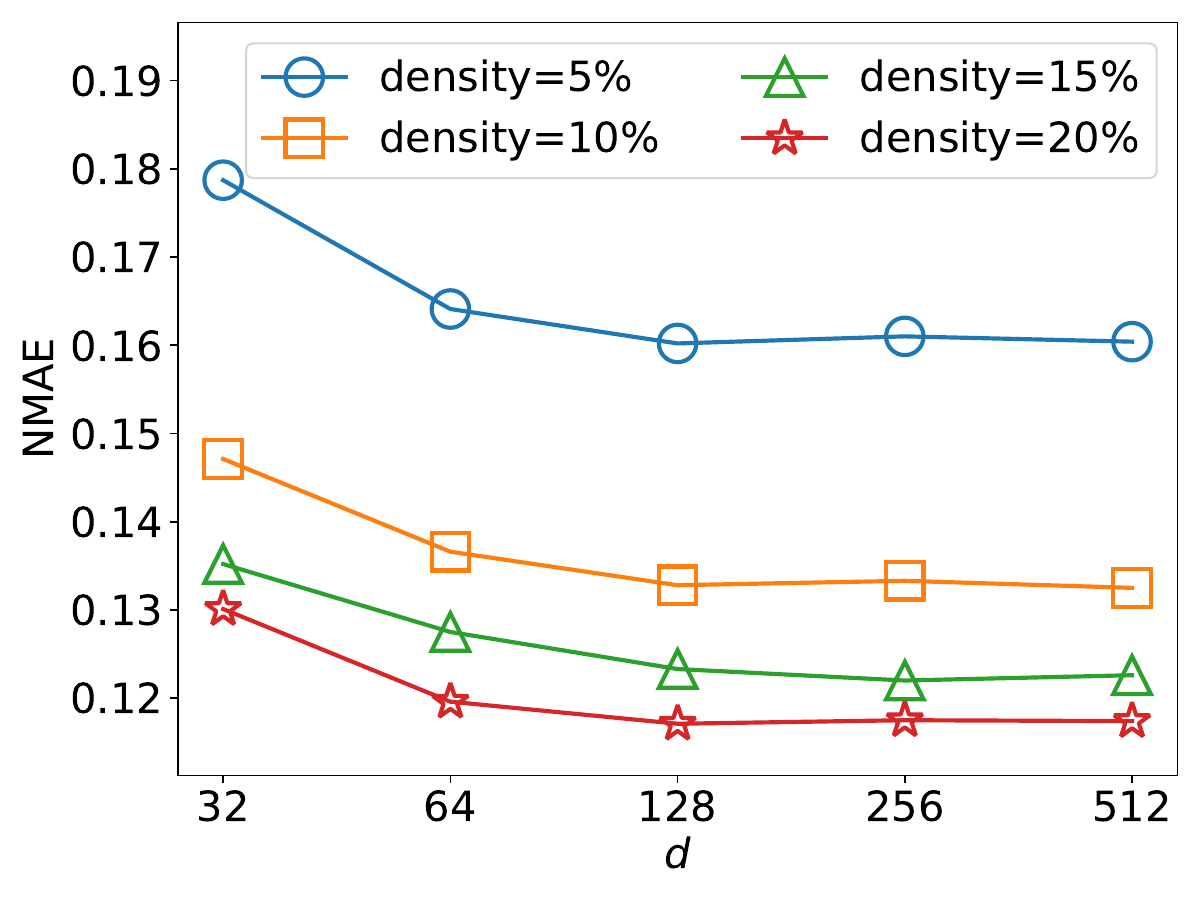}
        \label{fig:dim_rt_mae}
    }
    \hfill
    \subfloat[RMSE of various $d$ on RT]{%
        \includegraphics[width=0.23\textwidth]{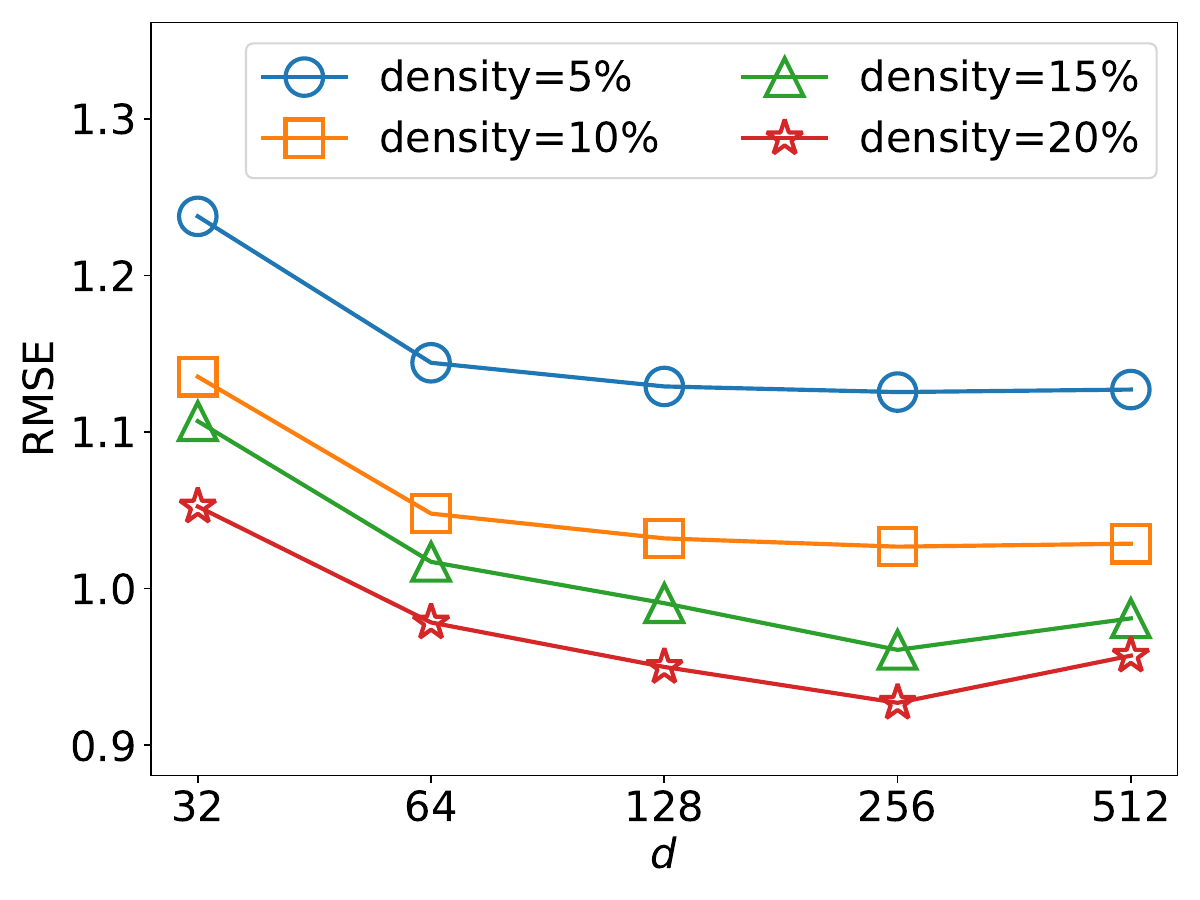}
        \label{fig:dim_rt_rmse}
    }
    \hfill
    \subfloat[NMAE of various $d$ on TP]{%
        \includegraphics[width=0.23\textwidth]{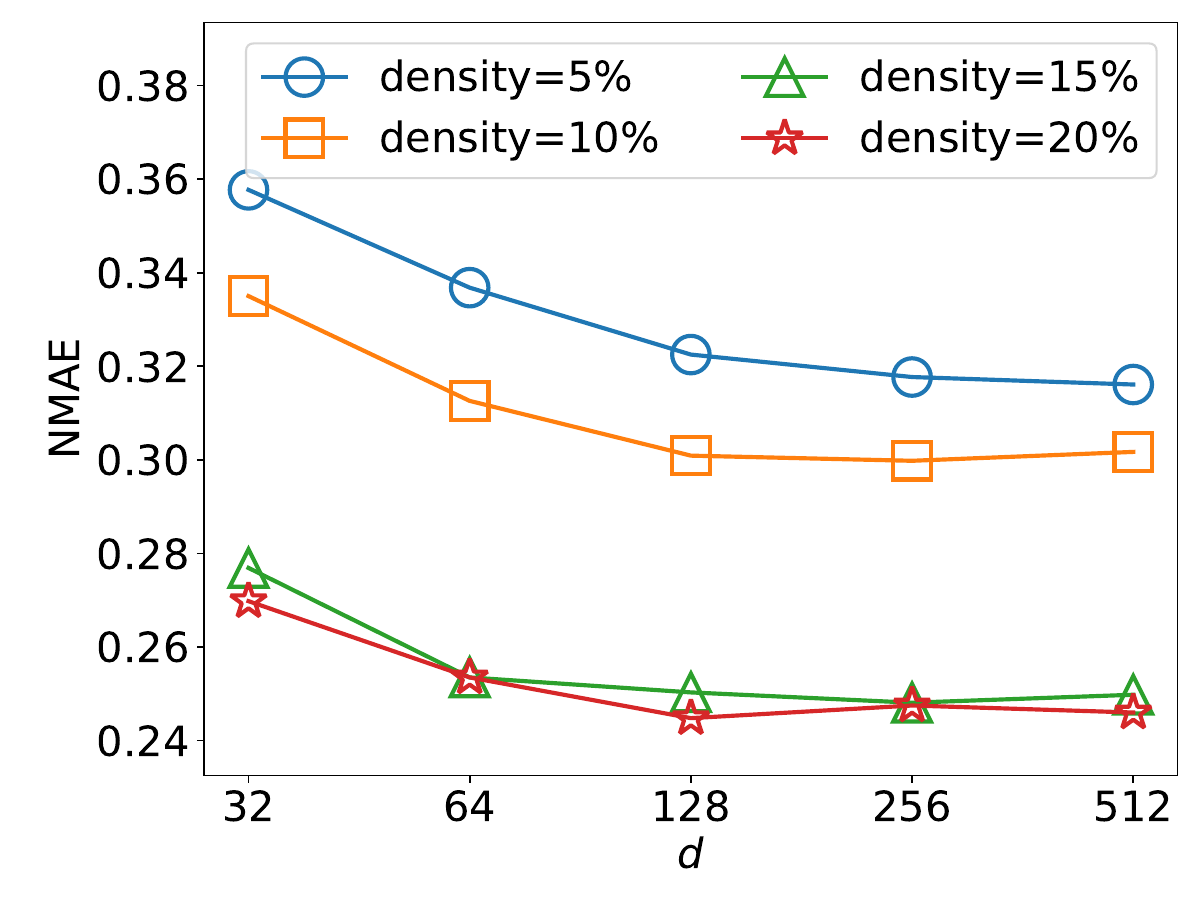}
        \label{fig:dim_tp_mae}
    }
    \hfill
    \subfloat[RMSE of various $d$ on TP]{%
        \includegraphics[width=0.23\textwidth]{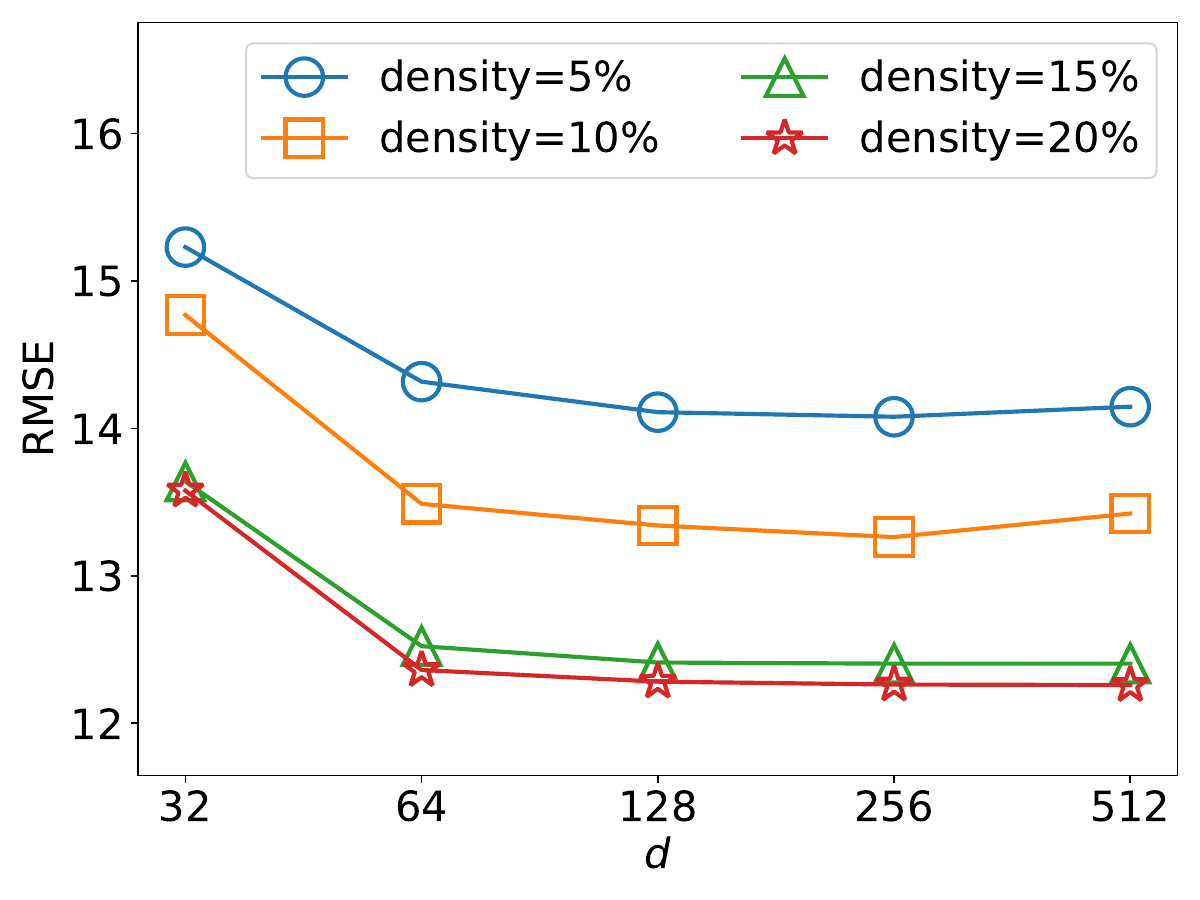}
        \label{fig:dim_tp_rmse}
    }
    \caption{Parameter impact of layers of target-prompt graph attention network $l_g$ and dimension of user/service feature $d$.}
    \label{fig:params}
\end{figure*}

Figure \ref{fig:ablations} illustrates the performance of the original GACL and its ablation variants. Each figure displays the metric values across different data density settings, allowing for a clear comparison of how each ablation variant impacts the model's performance. Figure \ref{fig:ab_rt_mae} shows the NMAE results on the RT dataset, while Figure \ref{fig:ab_rt_rmse} shows the RMSE results on the RT dataset. Similarly, Figure \ref{fig:ab_tp_mae} presents the NMAE results on the TP dataset, and Figure \ref{fig:ab_tp_rmse} presents the RMSE results on the TP dataset.

From the figures, we observe that on both the RT and TP datasets, the original GACL consistently achieves the lowest NMAE and RMSE across all QoS density settings, indicating its superior accuracy. The GACL-tw variant shows a noticeable increase in NMAE and RMSE, highlighting the importance of introducing the target-prompt module. The GACL-w and GACL-t variants also exhibit higher NMAE and RMSE values compared to GACL, but their performance is relatively better than GACL-tw, suggesting that historical QoS values and implicit collaborative relationships between target and neighbors play a significant role in reducing prediction errors.

Notably, on the RT dataset, GACL-w often achieves better performance than GACL-t, whereas on the TP dataset, the performances of GACL-t and GACL-w are comparable. This discrepancy is likely related to the different historical QoS distributions of the two datasets. The QoS value range of RT is much smaller compared to TP, with fewer outliers. For a specific user-service invocation, its QoS value and implicit relationship with the network context of target users and services can be more effectively mined by deep models. Therefore, in the target-prompt module, considering only the implicit correlation between target users/services and neighbors can achieve good QoS prediction accuracy. Conversely, for the TP dataset with a wider distribution range and more outliers, a specific user-service invocation's QoS value does not always consistently correlate implicitly with the network context of target users/services. Thus, it requires additional consideration of the impact of historical QoS values. As a result, both GACL-t and GACL-w experience significant performance losses compared to GACL.

In summary, the ablation studies demonstrate the effectiveness of the target-prompt attention strategy. Learning context-aware features of users and services for each distinct invocation by simultaneously considering implicit collaborative relationships, and historical QoS values significantly enhances the model's predictive accuracy and robustness. The superior performance of the original GACL across all metrics and datasets highlights the value of integrating these components in temporal QoS prediction.

\subsection{Performance Impact of Parameters}

In the experiments, five main hyperparameters significantly impact the performance of our proposed approach, including the layers of target-prompt graph attention network $l_{g}$, the dimension of high-order latent features of users/services $d$, window size $ws$, layers of transformer encoder $l_{tf}$ and head number of multi-head attention $l_{hd}$.
We conducted a series of ablation experiments on each parameter separately to verify the impacts of these hyperparameters for QoS prediction. As in the previous part, we conduct experiments on RT and TP datasets with four different QoS densities, and then report the performance on NMAE and RMSE respectively.

\subsubsection{\textbf{Performance Impact Analysis of $l_g$}}
The impacts of varying the number of layers $l_{g}$ are illustrated at Figure \ref{fig:hop_rt_mae} - \ref{fig:hop_tp_rmse}.
Figures \ref{fig:hop_rt_mae} and \ref{fig:hop_rt_rmse} show the NMAE and RMSE results on the RT dataset, while Figures \ref{fig:hop_tp_mae} and \ref{fig:hop_tp_rmse} present the corresponding metrics on the TP dataset, as $l_{g}$ varies from 1 to 4 hops. 
Various $l_{g}$ determines the extent of neighborhood information aggregated during the feature extraction process of target users and services. Specifically, a higher $l_{g}$ means that the model can capture more distant relationships in the graph, while a lower $l_{g}$ focuses on immediate neighbor users and services.

From the results, it is observed that both NMAE and RMSE achieve the best performance when $l_{g}$ is set to 2 or 3 hops across different data density settings. For instance, on the RT dataset, the NMAE decreases as $l_{g}$ increases from 1 to 2 hops, reaches a minimum at 2 hops, and slightly increases at 3 hops before rising more noticeably at 4 hops. A similar pattern is observed for RMSE on the RT dataset. On the TP dataset, the optimal performance is also seen at 2 or 3 hops, with a notable increase in both NMAE and RMSE when $l_{g}$ is set to 4 hops.

These observations can be attributed to several factors. 
When $l_{g}$ is set to 2 or 3 hops, the model effectively balances local and global information, capturing sufficient context from the directly interacted neighboring users or services while also incorporating relevant information from slightly more distant and indirectly interacted neighboring users or services. 
These close neighbors have intersecting invocation behaviors or environmental contexts with target user and service, all contribute to effectively extracting the characteristics of target users and services, thereby achieving accurate QoS prediction.
When $l_{g}$ is increased to 4 hops, the model starts to aggregate information from a larger neighborhood, which can lead to over-smoothing, reducing the model's ability to distinguish between different nodes and thus degrading performance. 
At 1 hop, the model only considers immediate neighbors, which might not provide enough context for accurate predictions, as reflected in the higher NMAE and RMSE values compared to 2 or 3 hops.

Overall, the optimal performance at 2 or 3 hops demonstrates that considering an intermediate range of neighbors allows the model to capture a rich set of features without the drawbacks of over-smoothing. 
\begin{figure*}[t]
    \centering
    \subfloat[NMAE of various $ws$ \& $l_{tf}$  on RT]{%
        \includegraphics[width=0.23\textwidth]{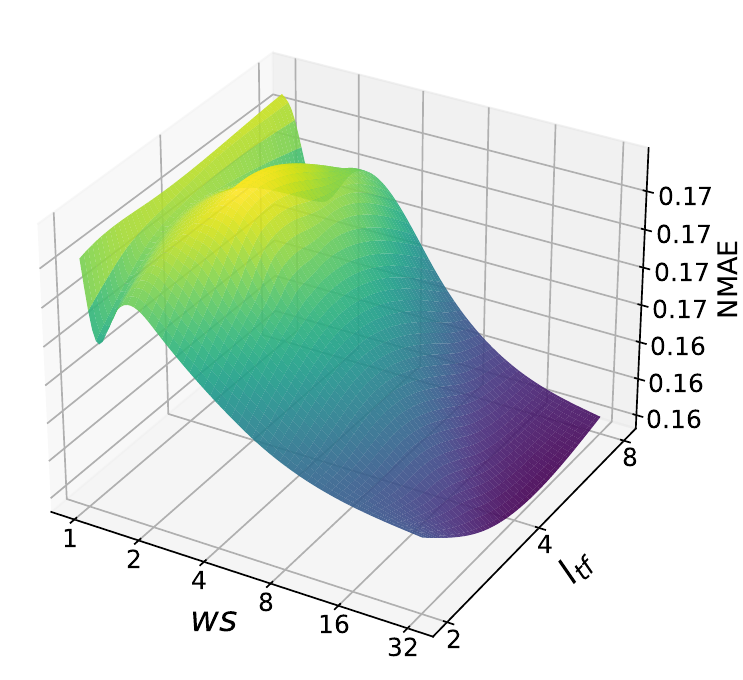}
        \label{fig:ws_rt_mae}
    }
    \hfill
    \subfloat[RMSE of various $ws$ \& $l_{tf}$ on RT]{%
        \includegraphics[width=0.23\textwidth]{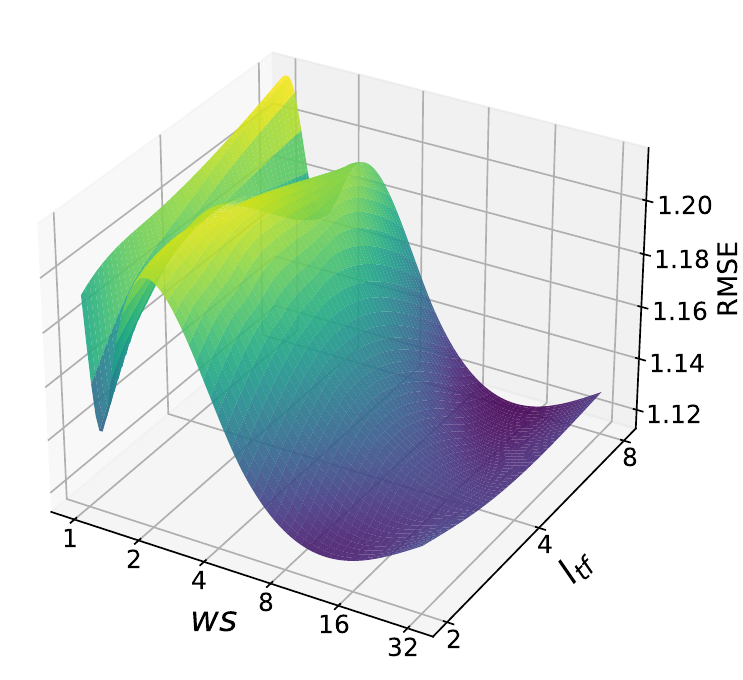}
        \label{fig:ws_rt_rmse}
    }
    \hfill
    \subfloat[NMAE of various $ws$ \& $l_{tf}$ on TP]{%
        \includegraphics[width=0.23\textwidth]{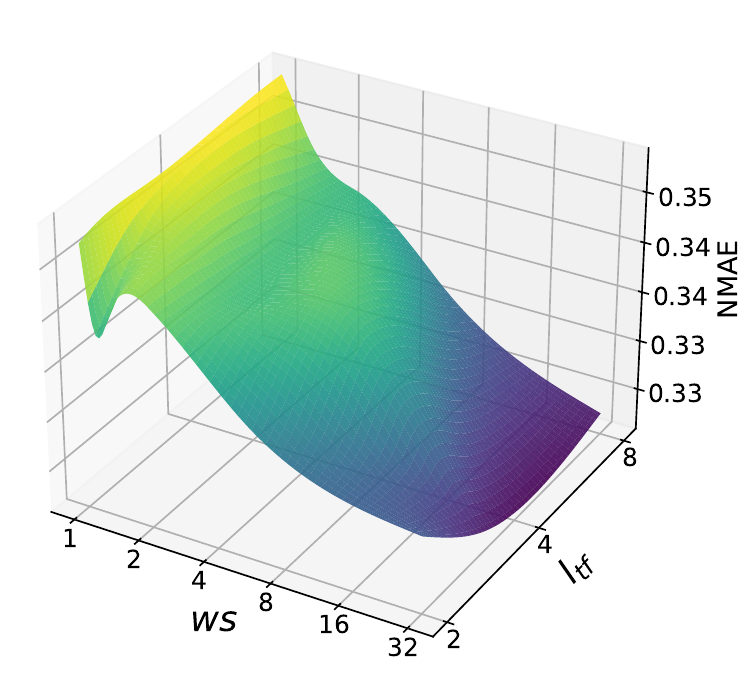}
        \label{fig:ws_tp_mae}
    }
    \hfill
    \subfloat[RMSE of various $ws$ \& $l_{tf}$ on TP]{%
        \includegraphics[width=0.23\textwidth]{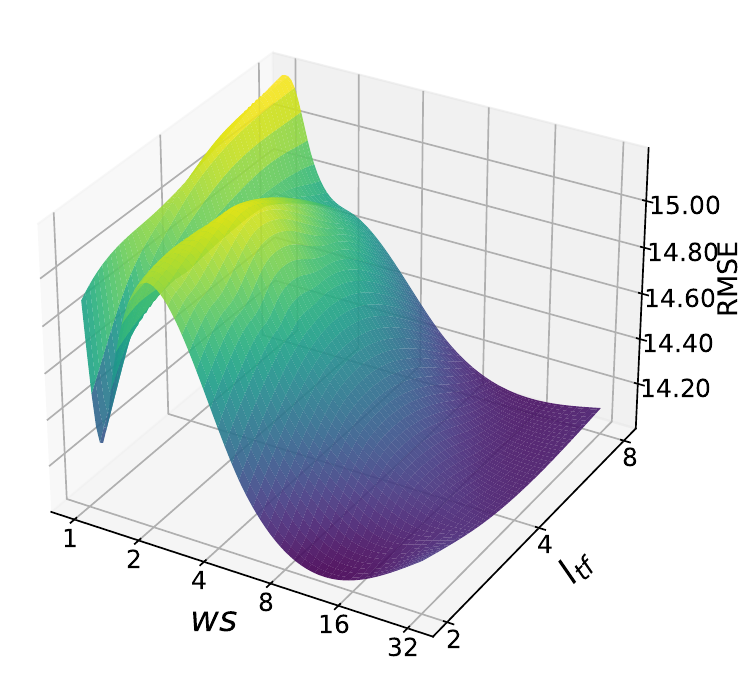}
        \label{fig:ws_tp_rmse}
    }
    \caption{Parameter impact of various combination of window size $ws$ and layers of the transformer encoder $l_{tf}$.}
    \label{fig:params_3d}
\end{figure*}
\subsubsection{\textbf{Performance Impact Analysis of $d$}}
The impacts of QoS prediction accuracy with different dimensionality $d$ are shown in Figure \ref{fig:dim_rt_mae}-\ref{fig:dim_tp_rmse}, as $d$ varies from 32 to 512. 
A larger $d$ allows for a richer and more expressive representation of user and service features, while a smaller $d$ limits the model's capacity to capture complex patterns.
From the results, it is evident that both NMAE and RMSE metrics generally improve as $d$ increases, with the best performance typically observed at $d=128$ or 
$d=256$ across different data density settings.
From the results, it is evident that both NMAE and RMSE metrics generally improve as $ d $ increases, with the best performance typically observed at $ d = 128 $ or $ d = 256 $ across different data density settings.
Specifically, the NMAE decreases significantly as $ d $ increases from 32 to 128. After $ d = 256 $, the improvement becomes more gradual.

These observations can be attributed to the following factors. When $ d $ is too small, the feature space is insufficient to capture the complex patterns and interactions between users and services, leading to higher prediction errors. As $ d $ increases, the model gains a richer and more expressive feature space, allowing it to better capture the nuanced relationships necessary for accurate QoS predictions. This results in a significant reduction in both NMAE and RMSE.
However, when $ d $ is increased beyond a certain point (e.g., 128 or 256), the performance gains diminish, and there may even be a slight degradation in performance. This can be due to overfitting, where the model becomes too complex and starts to capture noise in the training data, rather than the underlying patterns. Additionally, larger feature dimensions can lead to increased computational complexity and the risk of over-smoothing, where the features of different nodes become too similar.

Overall, the optimal performance at $ d = 128 $ or $ d = 256 $ demonstrates that a balanced feature dimension allows the model to capture a rich set of features without the drawbacks of overfitting. 

\subsubsection{\textbf{Performance Impact Analysis of Various Combination of $ws$ and $l_{tf}$}}
Figure \ref{fig:ws_rt_mae} - \ref{fig:ws_tp_rmse} show the impact of varying the window size ($ ws $) and the number of transformer encoder layers ($ l_{tf} $) on the QoS prediction performance. It is important to note that, for convenience, we set the number of multi-head attention heads ($ l_{h} $) equal to $ l_{tf} $.

From the results, it is evident that with the increase of $ws$ and $l_{tf}$, lower NMAE and RMSE can be achieved in both datasets and data density settings.
As $ ws $ increases, the model considers a broader temporal window, capturing long-term dependencies and global patterns in the user-service interactions. In this case, a larger number of transformer layers ($ l_{tf} $) is necessary to adequately model these long-term relationships and provide a comprehensive representation of the temporal dynamics. This is reflected in the improved performance at larger $ ws $ and larger $ l_{tf} $ settings. The increased $ ws $ helps to smooth out short-term fluctuations and capture stable long-term trends, which is crucial for accurate QoS prediction.

Moreover, the combination of smaller $ ws $ and smaller $ l_{tf} $ generally yields better performance. Similarly, as $ ws $ increases, larger $ l_{tf} $ tends to produce better results.
When $ ws $ is small, the model focuses on a narrow temporal window, capturing short-term dependencies and local patterns in the user-service interactions. In such scenarios, a smaller number of transformer layers is sufficient to model these temporal relationships effectively, while large $l_{tf}$ can easily lead to overfitting.
As $ ws $ increases, the model considers a broader temporal window, capturing long-term dependencies and global patterns in the user-service interactions. In this case, a larger number of transformer layers is necessary to adequately model these long-term relationships and provide a comprehensive representation of the temporal dynamics. This is reflected in the improved performance at larger $ ws $ and larger $ l_{tf} $ settings.

This analysis underscores the importance of tuning both $ ws $ and $ l_{tf} $ to match the temporal characteristics of the user-service interactions being modeled. So in our experiments scenerios, we set $ws=32$ and $l_{tf}=l_{h}=8$.

%% file: related_work.tex
\section{Related Work}
\label{sec:related_work}

\subsection{Static QoS Prediction}
Static QoS prediction approaches can be classified into three categories: memory-based, model-based, and deep learning-based. These methods typically operate on a two-dimensional matrix representing user-service QoS invocations.

Memory-based methods primarily utilize traditional collaborative filtering (CF) techniques to predict missing QoS values. These approaches can be further divided into user-based \cite{shao2007personalized}, service-based \cite{sarwar2001item}, and hybrid methods that combine both user-based and service-based predictions using weighted coefficients. The fundamental principle of memory-based QoS prediction is to identify a set of similar users or services (neighborhood) through similarity calculations and then use this neighborhood to compute deviation migrations. These migrations are combined with average QoS values to predict the missing QoS. Various studies have focused on improving the similarity quantification between users and services to enhance neighborhood recognition \cite{wang2017hybrid}.
Wu et al. \cite{wu2015collaborative} introduced a rate-based similarity (RBS) method to select user and service neighborhoods, achieving better QoS predictions.
Zou et al. \cite{zou2018qos} proposed a reinforced CF approach that combines RBS and Pearson Correlation Coefficient (PCC) to accurately calculate average QoS values and deviation migrations.

Model-based approaches aim to extract implicit linear or nonlinear invocation relationships to enhance QoS prediction performance, partially addressing the limitations of CF-based methods. Xu et al. \cite{xu2016context} proposed two context-aware matrix factorization models that improve QoS prediction by considering user and service contexts. Wu et al. \cite{wu2018collaborative} introduced a general context-sensitive matrix factorization approach to model interactions between users and services more effectively.

Deep learning techniques have recently been employed to solve QoS prediction problems due to their ability to handle data sparsity and learn implicit nonlinear interactions. These methods often combine neural networks with matrix factorization and adopt multi-task learning to reduce prediction errors and improve performance. For instance, Xu et al. \cite{xu2021nfmf} developed the model that integrate deep learning with matrix factorization to enhance prediction accuracy. Li et al. \cite{li2021topology} proposed the topology-aware neural (TAN) model, which considers the underlying network topology and complex interactions between autonomous systems to improve collaborative QoS prediction. Zou et al. \cite{zou2023ncrl} designed a location-aware two-tower deep residual network combined with collaborative filtering, achieving superior QoS prediction. Recent advancements have further improved QoS prediction performance by incorporating expert systems and attention mechanisms \cite{lian2023toward}, or using GNNs \cite{wu2023robust} to select, extract, and interact with multiple features from user-service contextual information and QoS invocations.

\subsection{Temporal QoS Prediction}
Temporal QoS prediction can be partitioned into four categories: temporal factor integrated CF, sequence prediction, tensor decomposition, and deep learning.

Temporal factor integrated CF methods incorporate temporal information into the collaborative filtering process. Hu et al. \cite{hu2014time} integrated temporal factors with the CF approach and used a random walk algorithm to select more similar neighbors, alleviating data sparsity and improving temporal QoS prediction. Ma et al. \cite{ma2017multi} proposed a vector comparison approach combining orientation and dimension similarity to implement time series analysis for multi-valued collaborative QoS prediction in cloud computing. Tong et al. \cite{tong2021missing} improved temporal QoS prediction by normalizing historical QoS values, calculating similarity, and selecting neighbors based on the distance of time slices, then using hybrid CF for prediction. These approaches demonstrate the effectiveness of integrating temporal information into QoS prediction and addressing the limitations of non-temporal QoS prediction approaches.

Sequence prediction methods use time series analysis techniques to enhance temporal QoS prediction. Hu et al. \cite{hu2015web} combined CF with the ARIMA model and applied the Kalman filtering algorithm to compensate for ARIMA's shortcomings in temporal QoS prediction. Ding et al. \cite{ding2018time} integrated the ARIMA model with memory-based CF to capture the temporal characteristics of user similarity, improving the accuracy of missing temporal QoS predictions. These approaches highlight the benefits of combining sequence prediction analysis with QoS prediction to capture temporal characteristics of QoS variations.

Tensor decomposition methods convert the classic two-dimensional user-service matrix into a three-dimensional tensor representation, enabling temporal factor incorporation. Meng et al. \cite{meng2016temporal} proposed a temporal hybrid collaborative cloud service recommendation approach using CP decomposition and a biases model to distinguish temporal QoS metrics from stable ones. Zhang et al. \cite{zhang2019recurrent} combined Personalized Gated Recurrent Unit (PGRU) and Generalized Tensor Factorization (GTF) to leverage long-term dependency patterns for comprehensive temporal QoS prediction. Luo et al. \cite{luo2019temporal} introduced a temporal pattern-aware QoS prediction approach using a biased non-negative latent factorization of tensors (BNLFTs) model to extract time potential factors from dynamic QoS. These methods show the effectiveness of using tensor representations and factorization techniques to incorporate temporal factors into QoS prediction.

Deep learning models, such as RNN and its variants LSTM and GRU, have been increasingly used for temporal QoS prediction. Xiong et al. \cite{xiong2017learning} considered multi-dimensional context to develop an effective QoS prediction model from past QoS invocation history. Xiong et al. \cite{xiong2018personalized} proposed a personalized matrix factorization approach (PLMF) based on LSTM to capture dynamic representations for online QoS prediction. Zou et al. \cite{zou2022deeptsqp} developed a temporal QoS prediction framework called DeepTSQP, which combines binary features with memory-based similarity and feeds them to a GRU model to mine temporal aggregated features for predicting unknown temporal QoS values. These deep learning approaches effectively capture temporal dependencies and patterns, enhancing the accuracy of temporal QoS prediction.

In summary, the existing research on QoS prediction has made significant strides in both static and temporal contexts. Building on these advancements, we introduce a novel GACL framework that leverages a dynamic user-service invocation graph and develops a target-prompt graph attention network to extract invocation-specific deep latent features of users and services. By incorporating temporal evolution patterns using a multi-layer Transformer encoder, our approach addresses the limitations of existing methods and achieves more precise temporal QoS predictions. This development of advanced graph neural networks and the integration of temporal modeling techniques represents a significant contribution to the field, providing a robust solution for accurate QoS prediction in dynamic service-oriented environments.